\documentclass{article}

\usepackage[square,numbers]{natbib}
\usepackage[final]{neurips_2024}

\usepackage[utf8]{inputenc} 
\usepackage[T1]{fontenc}    
\usepackage[citecolor=blue, colorlinks]{hyperref}
\usepackage{url}            
\usepackage{booktabs}       
\usepackage{amsfonts}       
\usepackage{nicefrac}       
\usepackage{microtype}      
\usepackage{xcolor}         
\usepackage{graphicx}   
\usepackage{algorithm}
\usepackage{algorithmicx}
\usepackage{mathtools}
\usepackage{amsmath}
\usepackage{array}
\usepackage{subcaption}

\title{E-Motion: Future Motion Simulation via Event Sequence Diffusion}

\author{
  Song Wu $^1$, Zhiyu Zhu $^2$, Junhui Hou $^2$, Guangming Shi $^1$ , Jinjian Wu $^{1}$ \thanks{This work was supported in part by NSFC Excellent Young Scientists Fund 62422118. The first two authors contributed to this paper equally.   Corresponding author: Jinjian Wu}  \\
  $^1$ Xidian University, $^2$ City University of Hong Kong \\ 
  \texttt{swu\_666@stu.xidian.edu.cn,zhiyuzhu2-c@my.cityu.edu.hk} 
  \\
  \texttt{jh.hou@cityu.edu.hk,gmshi@xidian.edu.cn,jinjian.wu@mail.xidian.edu.cn} 
}

\begin{document}

\maketitle

\begin{abstract}
Forecasting a typical object's future motion is a critical task for interpreting and interacting with dynamic environments in computer vision. Event-based sensors, which could capture changes in the scene with exceptional temporal granularity, may potentially offer a unique opportunity to predict future motion with a level of detail and precision previously unachievable. Inspired by that, we propose to integrate the strong learning capacity of the video diffusion model with the rich motion information of an event camera as a motion simulation framework. Specifically, we initially employ pre-trained stable video diffusion models to adapt the event sequence dataset. This process facilitates the transfer of extensive knowledge from RGB videos to an event-centric domain. Moreover, we introduce an alignment mechanism that utilizes reinforcement learning techniques to enhance the reverse generation trajectory of the diffusion model, ensuring improved performance and accuracy. Through extensive testing and validation, we demonstrate the effectiveness of our method in various complex scenarios, showcasing its potential to revolutionize motion flow prediction in computer vision applications such as autonomous vehicle guidance, robotic navigation, and interactive media. Our findings suggest a promising direction for future research in enhancing the interpretative power and predictive accuracy of computer vision systems. The source code is publicly available at \url{https://github.com/p4r4mount/E-Motion}.

\end{abstract}

\section{Introduction}
\label{sec:intro}

Accurately capturing and interpreting dynamic scenes under fluctuating motion and illumination conditions remains an enduring challenge in computer vision~\cite{liang2020learning,nayakanti2023wayformer}. This challenge is particularly pronounced in real-world settings, where subtle variations can dramatically affect the perception and analysis of future motion~\cite{xu2024towards,ettinger2021large}. Traditional imaging modalities often struggle to capture these nuances, leading to a gap in accurately modeling and predicting motion flow in complex visual environments.

The rapid advancements in deep learning have catalyzed transformative developments in computer vision, particularly in the generative models~\cite{sohl2015deep, goodfellow2020generative, ho2020denoising, kingma2021variational, song2020denoising, song2020score,song2023consistency}. Video diffusion models~\cite{blattmann2023stable, ho2022imagen,mei2023vidm,esser2023structure,blattmann2023videoldm}, which stand at the forefront of these innovations, leverage stochastic diffusion processes to generate, restore, and accurately manipulate video content. These models, emblematic of the state-of-the-art in temporal data processing, offer refined capabilities for complex video-based tasks~\cite{chai2023stablevideo,ruan2023mm}, underscoring the significant strides made in understanding and interpreting dynamic visual scenes.

Coming into the high temporal resolution field, event data stands as a revolutionary sensing approach~\cite{rebecq2019high,gallego2020event,tulyakov2021time} that significantly mitigates this gap and consistently captures even very subtle fluctuations. Such strong capacity comes from the unique sensing pattern of the event camera, which asynchronously measures the intensity variation in high temporal resolution. It allows for the precise detection of miniature illumination changes, providing a rich, granular record that traditional cameras simply cannot offer~\cite{zhu2022learning,zhu2023cross,chen2023segment}.

\begin{figure}[t]
    \centering
    \resizebox{0.8\textwidth}{!}{\includegraphics{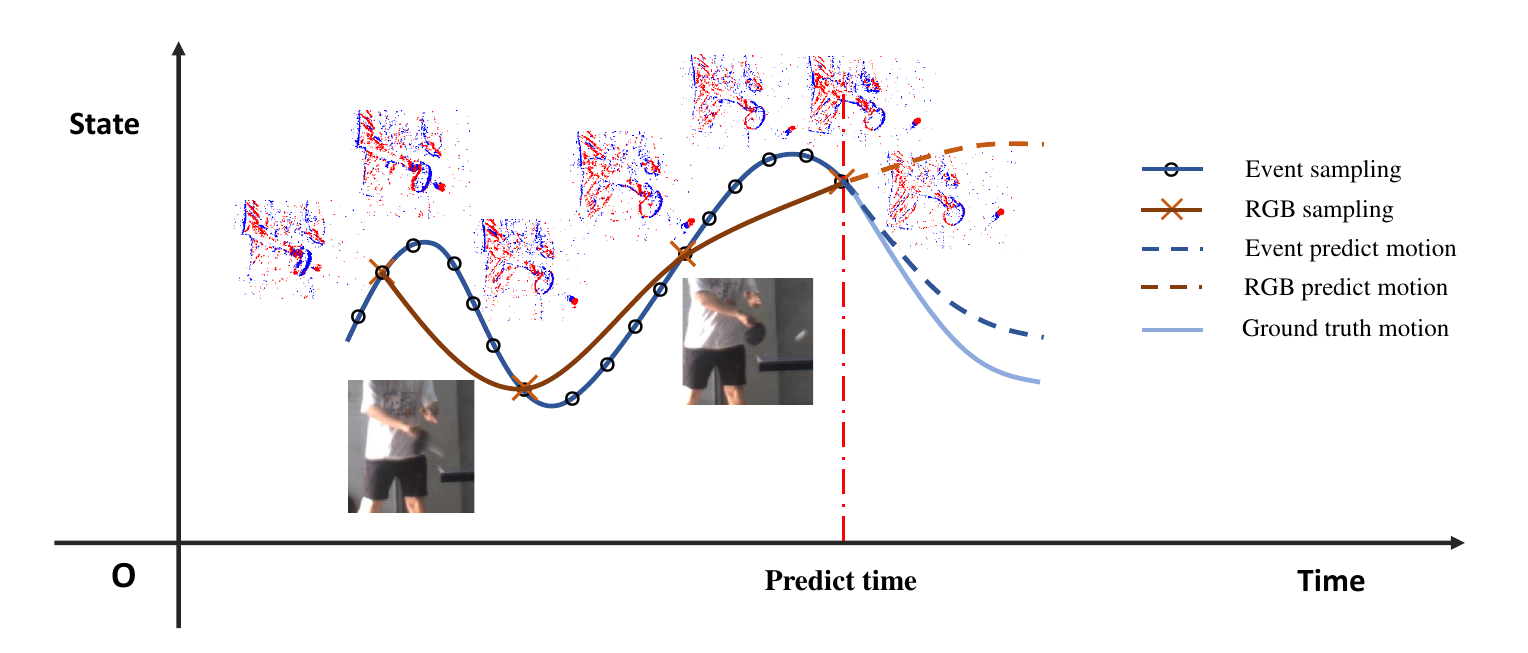}}
    \caption{Illustration that the exceptional temporal resolution afforded by event cameras, alongside their distinctive event-driven sensing paradigm, presents a significant opportunity for advancing the precision in predicting future motion trajectories.}
    \label{Fig:compare}
    \vspace{-0.4cm}
\end{figure}

To enable the video diffusion model to completely and correctly learn concise motion information for the estimation of potential future movement, integrating it with event data to realize an event sequence generation model is a potential solution that may inject high-frequency motion information into the video diffusion model. This paper delves into the symbiosis of video diffusion models and high-temporal-resolution event data, exploring its potential to redefine motion forecasting in computer vision. We commence by delineating the landscape of video diffusion models and the mechanics of event-based sensing, elucidating their complementary strengths. Subsequently, we introduce a novel framework that amalgamates these technologies, aiming to augment the precision of motion flow predictions. Through comprehensive experimentation, we validate the effectiveness of our methodology, demonstrating its superior performance across diverse scenarios. Our findings illuminate the path forward, showcasing the integration of video diffusion models with event data as a robust, innovative solution for capturing and interpreting the complexities of dynamic scenes with unparalleled detail and accuracy.

In summary, the contribution of this paper lies in the following three parts:
\begin{itemize}
    \item we make the first attempt to integrate event-sequences with a video diffusion model, resulting in an event-sequence diffusion model, which could potentially estimate future object motion, given by a certain event prompt;
    \item we propose to align the pre-trained event-sequence diffusion model with the real-world motion via a reinforcement learning process, which stabilizes the results generated by the diffusion model and makes them more closely resemble real motion.
    \item we integrate a test-time prompt augmentation method to make use of high temporal resolution event sequence prompt to enhance the generation performance.
\end{itemize}

\section{Related Work}

\subsection{Event-based Vision}
Event-based vision represents a paradigm shift from traditional frame-based imaging, offering a dynamic and highly granular approach to capturing visual information. Unlike conventional cameras that record static frames at fixed intervals, the event-based sensor~(developed by Lichtsteiner \textit{et al.}~\cite{4444573} and further elaborated by Posch \textit{et al.}~\cite{5648367}) asynchronously records intensity changes per pixel and generates a signal termed an ”event” whenever the intensity surpasses a threshold, thereby providing a continuous stream of data that reflects temporal changes with remarkable precision. This method is particularly effective in environments with rapid motion or varying illumination, where traditional cameras suffer from motion blur and latency issues.

Recent advancements in this field have focused on leveraging the high temporal resolution of event data for various applications, including high-speed tracking~\cite{10284004,mahlknecht2022exploring,zhao2022moving}, dynamic scene reconstruction~\cite{shaw2022hdr,jiang2023event,zouLearningReconstructHigh2021}, and optical flow estimation~\cite{shiba2022secrets,ding2022spatio,hu2022optical}. Works by Gallego \textit{et al.}~\cite{8578666} and Rebecq \textit{et al.}~\cite{8946715} have been instrumental in demonstrating the utility of event-based data in reconstructing high-speed phenomena and enhancing motion analysis, setting a solid foundation for our research.

\subsection{Multimodal Diffusion Models}
Generative diffusion models, introduced by Sohl-Dickstein \textit{et al.}~\cite{sohl2015deep}, represent a class of probabilistic generative models that simulate the gradual transformation of data from a complex distribution into a simpler, typically Gaussian distribution, and vice versa. ~\cite{croitoru2023diffusion,kingma2021variational,yang2023diffusion,cao2024survey}. This process, characterized by a series of forward and reverse diffusion steps, has been applied successfully to a range of tasks, including image synthesis~\cite{ho2022cascaded,nichol2021glide}, restoration~\cite{lin2023diffbir,hou2024global}, and, more recently, temporal data manipulation~\cite{mei2023vidm,ge2023preserve}.

The application of diffusion models to video data, as explored by Ho \textit{et al.}~\cite{ho2022imagen} and extended by others, marks a significant advancement in the field, offering new pathways for the generation and manipulation of dynamic scenes. These models have shown exceptional promise in capturing the temporal continuity and complexity inherent in video data, providing a robust framework for tasks such as video prediction and temporal interpolation.

In recent years, the development of multimodal diffusion technology has advanced rapidly. Researchers are dedicated to applying the powerful generative capabilities of diffusion to different modalities with unique advantages, such as optical flow and depth. Saxena et al.~\cite{saxena2024surprising} was the first to apply diffusion models to optical flow and depth estimation. For the characteristics of training data, they introduced infilling, step-rolling, and L1 loss during training to mitigate distribution shifts between training and inference. To address the lack of ground truth in datasets, they also used a large amount of synthetic data for self-supervised pretraining, enabling the diffusion model to acquire reliable knowledge. Chen et al.~\cite{chen2023controlavideo} utilized the motion information embedded in control signals such as edges and depth maps to achieve more precise control over the text-to-video (T2V) process. They used pixel residuals and optical flow to extract motion-prior information to ensure continuity in video generation. Additionally, they proposed a first-frame generator to integrate semantic information from text and images. 

Despite the extensive body of research within the domain of temporal analysis, it is important to acknowledge that the majority of these studies focus primarily on the domain of RGB images and videos. As previously discussed, the superior temporal resolution offered by event data holds significant potential for enhancing the alignment process. Consequently, it is imperative to undertake a thorough investigation into the application and adaptation of existing pre-trained diffusion models to the realm of event data.

\section{Preliminary}
\textbf{Diffusion Models} are a class of generative models that simulate the gradual transformation of data from a complex, high-dimensional distribution to a simpler, typically Gaussian distribution through a process known as forward diffusion~\cite{song2020score,ho2020denoising,song2020denoising,song2019generative}. Conversely, the reverse diffusion process aims to reconstruct the original data distribution from the simpler one. This mechanism is inspired by thermodynamic processes and has been increasingly applied in the field of deep learning for generating high-quality, diverse samples from complex distributions.

The mathematical foundation of diffusion models is rooted in stochastic differential equations (SDEs), which describe the forward and reverse diffusion processes. 
By score-based formulation~\cite{song2020score}, the forward process of diffusion model acts as
\begin{equation}
\label{eq:foward}
    d\textbf{x} = \textbf{f}(\textbf{x},t)dt + g(t)d\textbf{w},
\end{equation}
where $\textbf{x}\in \mathbb{R}^n$ indicates the state of reconstructed signal, $\textbf{w}\in \mathbb{R}^n$ represents a standard Wiener process ($g(t)d\textbf{w}\sim\mathcal{N}(0,g(t)^2\mathbf{I}d\textbf{w})$), $\textbf{f}(\cdot,t):\mathbb{R}^n \rightarrow \mathbb{R}^n$ and $g(\cdot):\mathbb{R} \rightarrow \mathbb{R}$ represent the drift and diffusion coefficients, respectively.
Moreover, in the evaluating phase, the reverse~(inference) process could be illustrated as iteratively performing the following ODE step:
\begin{equation}
    d\textbf{x} = [\textbf{f}(\textbf{x},t) - \frac{1}{2}g(t)^2\nabla_\textbf{x} log p(\textbf{x})] dt,
\end{equation}
where $\nabla_\textbf{x} log p(\textbf{x})$ is usually approximated by a learnable score model $S_{\theta}(\mathbf{x},t)$. Based on the theoretical formulation of stable video diffusion~\cite{blattmann2023stable}, e.g., variance exploding (VE) diffusion process~\cite{song2020score}, the reverse process is acted as 
\begin{equation}
\label{eq:DDIM}
    \textbf{x}_{t-1} = \textbf{x}_{t} - \frac{\textbf{x}_{t}-\mu_\theta(\textbf{x}_{t},t)}{\sigma_t}(\sigma_{t}-\sigma_{t-1}),
\end{equation}
where $\mu_\theta(\cdot)$ is one of parametrization method of $S_{\theta}(\cdot)$, which estimate the clean image from noise latent $\textbf{x}_{t} \sim \mathcal{N}(\mathbf{x}_0,\sigma_t^2\mathbf{I})$.

\begin{figure}[t]
    \centering
        \resizebox{1.0\textwidth}{!}{\includegraphics{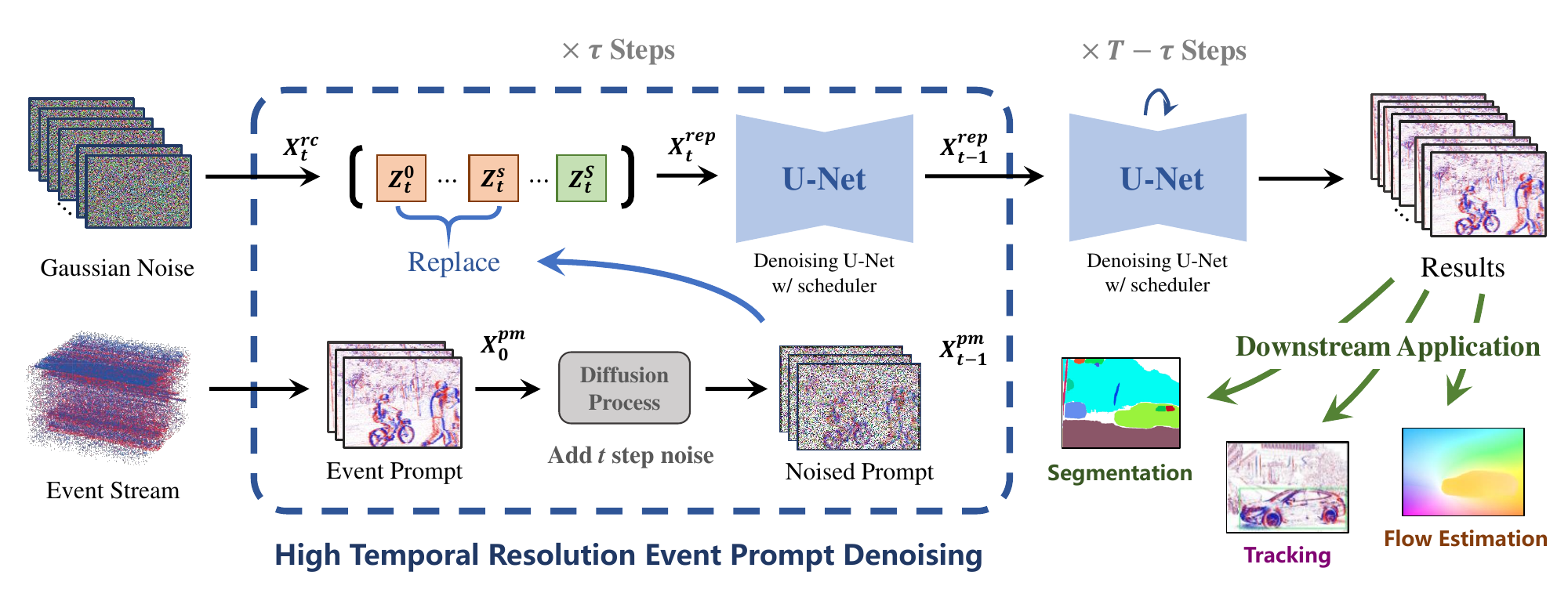}}
    \caption{Inference workflow of the proposed method, where the left upper one indicates the random Gaussian noise, left lower one represents the prompted event sequence. We perform $\tau$ steps forward diffusion processing on the event prompt and substitute a portion of the diffusion input noise, followed by $T-\tau$ Steps of conventional denoising.}
    \label{Fig:Diff}
    \vspace{-0.2cm}
\end{figure}

\section{Methods}

The endeavor of video diffusion presents multifaceted challenges, notably the simultaneous generation of sequences that not only manifest dynamic, convincing motion but also maintain authentic textures. The utilization of event data, characterized by its intrinsic high temporal resolution, emerges as a strategic solution for the precise modeling of object motion. In contrast to traditional stable video diffusion models, which are typically initialized using a singular image or textual prompt, our proposed methodology, designated as the Event-Sequence Diffusion Network, capitalizes on a succinct sequence of events as its conditioning input. This novel approach is illustrated in Fig.~\ref{Fig:Diff}.

As elaborated in Sec.~\ref{Sec:Pretrain}, we delineate the comprehensive pre-training regimen that facilitates the learning of object motion through the prediction of subsequent events. Nevertheless, due to the intrinsic diversity-generating property of diffusion models, they commonly yield several distinct samples from a single input prompt. Although these variations may each seem feasible, there is no assurance that they align consistently with the actual dynamics of real-world motion. Thus, in Sec.~\ref{Sec:aligning}, we augment the quality of generation by incorporating an alignment process. It employs reinforcement learning techniques to impose a structured regularization on the generative process of the diffusion models, thereby enhancing the fidelity and coherence of the produced sequences.

\subsection{Learning Motion Prior via Pretraining on Event Sequences}
\label{Sec:Pretrain}
 

An initial formatting process is discussed to retain its high-temporal resolution properties effectively and to facilitate integration with pre-existing large-scale models. Furthermore, to mitigate the substantial computational demands associated with video diffusion models, we have implemented a prompt sampler designed to enhance the efficiency of information encapsulation derived from event sequence diffusion frameworks. Subsequently, we will delineate these critical aspects in a detailed manner.

\noindent \textbf{Event Representation.} 
To leverage contemporary video diffusion models for event data generation, we adopt a strategy where both event information and corresponding images are concurrently inputted into the video diffusion framework, which then processes adaptively sampled outcomes. Specifically, as illustrated in Fig.~\ref{Fig:Diff}, for a given event denoted as $\mathbf{E} = \{ h, w, p, t \}$, we consolidate the event data into a voxel grid representation~\cite{zhu2018ev}, symbolized as $\Bar{\mathbf{E}} \in \mathbb{R}^{B\times H\times W}$, where B denotes the number of time bins. In order to utilize the rich pre-training information from RGB frames, we set $B=3$. This approach ensures that the video diffusion model can effectively interpret and integrate the high-dimensional event data alongside conventional image inputs, facilitating a more comprehensive synthesis of dynamic visual content.

\noindent \textbf{Pretraining.}  
The training regimen for our proposed approach is similar with that of diffusion models ~\cite{song2020score,karras2022elucidating}. This methodology involves estimating the clean image from perturbed samples, as
\begin{equation}
    \mathcal{L}_{pre} = \mathbb{E}_t\{\lambda(t) \mathbb{E}_{\textbf{x}_{0}}\mathbb{E}_{\textbf{x}_{t}|\textbf{x}_{0}}[\| x_0 - \mu_\theta(\textbf{x}_{t},t) \|_2^2]\},
\end{equation}
where $\lambda(t)$ denotes a weighting function, parameterized as $\frac{\sigma_t^2 + \sigma_{data}^2}{(\sigma_t + \sigma_{data})^2}$
During training, we randomly select a set of intermediate states $\mathbf{X}_{t}$ and apply regularization techniques to guide them towards the accurate estimation of the underlying noise component $\epsilon$. Moreover, to make the diffusion network adaptively capture the object motion with different time windows, in the training process we randomly augment the diffusion network with the voxel from different time ranges. (\textit{Please refer to the Appendix Sec.~\ref{sec:details} for more details})

\noindent \textbf{High-temporal Resolution Guided Sampling.}~Owing to the unique operational mechanism of event cameras, which detect changes in intensity rather than capturing static images, the clarity and definition of recorded objects can significantly diminish if the chosen temporal window is either too brief or excessively prolonged, especially when we only feed one prompt, like a standard stable video diffusion network.
This characteristic often results in blurred or indistinct imagery when the temporal resolution does not align optimally with the scene's dynamics.

Inspired by the recent advancement of guided sampling~\cite{pan2023accurate}, we propose to aggregate multiple high-temporal resolution event frames as a test-time prompt. Specifically, as illustrated in Fig.~\ref{Fig:Diff}, during the inference process, we feed $s>1$ frames instead of a single frame as prompt $\mathbf{X}^{pm}_{0} \in \mathcal{R}^{s\times h \times w \times 3}$. For the step $T$, both $\mathbf{X}^{pm}_{T}$ and $\mathbf{X}^{rc}_{T}$ are initialized with the random Gaussian noise. However, for the step $t < \tau$, we set $\mathbf{X}^{pm}_{t-1}\leftarrow \mathbf{X}_0^{pm} + \sigma_t\epsilon$, Subsequently, the noised event prompt $\mathbf{X}^{pm}_{t-1}$ replaces the first $s$ random noises in the noise tensor $\mathbf{X}^{rc}_{t}$, providing motion priors for the denoising process. (\textit{Please refer to the Appendix Algorithm~\ref{alg2} for more details})

\begin{algorithm}[t]
  \caption{Motion Alignment Process}
  \label{alg1}
  \begin{algorithmic}[1]
    \vspace{1.2mm}
    \State \textbf{Input}: Reference model $\theta'$,  training model $\theta$ and reward model $\mathcal{R}(\mathbf{x})$.
    \State \textbf{For} $i=1,...,I$ \textbf{do}
    \State ~~~~Sample reference trajectory $\{x_O| \pi_{\theta'}\} , \mathcal{R}(\mathbf{x}_O)$ and $P_{\theta'}$.
    \State~~~~\textbf{For} $t=1,...,T$ \textbf{do}
    \State~~~~~~~~Calculate $P_{\theta}$ based on reference trajectory $\{x_O| \pi_{\theta'}\} $.
    \State~~~~~~~~Optimizing $\nabla_{\theta} \mathcal{J}'(\mathbf{\mathbf{x}})$.
    \State ~~~~\textbf{End for}
    \State ~~~~$\theta' \leftarrow \theta$.
    \State \textbf{End for}
    \State \textbf{Return} $\theta$
    \vspace{1.2mm}
  \end{algorithmic}
\end{algorithm}

\subsection{Motion Alignment via Reinforcement Learning}
\label{Sec:aligning}
Given the inherent challenges in precisely tailoring diffusion models to fit the entirety of the training data, particularly when considering the disparity between the model's size and the volume of the dataset, it is pragmatic to direct potential losses towards regions less perceptible to end-users or higher-level algorithms. Furthermore, the multi-step nature of the diffusion generation process renders the simultaneous training of the entire pipeline nearly unfeasible. To address this, we employ a strategy of reinforced optimization, conceptualizing the generation process as a Markov chain. This approach underscores the importance of guiding the diffusion process to yield results of superior quality, thereby optimizing the model's performance while accommodating its structural and computational complexities. Optimizing such a diffusion model starts with the following equations:
\begin{equation}
    \max_{\theta} \hspace{0.25em} \mathcal{J} = \int_{\mathbf{P}_\theta(\mathbf{x})} \mathcal{R}(\mathbf{x}) d\mathbf{x},
\end{equation}
where $\mathbf{P}_\theta(\mathbf{x})$ indicates the distribution of reconstructed samples under the model weight $\theta$. Although the diffusion model is a probabilistic model, its weights are generally deterministic. ~\cite{song2020denoising} The randomness generally comes from Gaussian sampling. Thus, we adopt the same measurement as~\cite{zhang2024large} to utilize the Gaussian density function $\mathcal{N}(\mathbf{x}_t|\mu_t,\sigma) \sim \frac{1}{\sigma\sqrt{2\pi}} e^{-\frac{(\mathbf{x}_t-\mu(\mathbf{x}_t, t))^2}{2\sigma^2}}$ to measure the $\mathbf{P}_\theta(\mathbf{x})$ of a given sample $\mathbf{x}_t$, where $\mu(\cdot)$ indicates the estimated clean latent without adding random noise. We then adopt the policy gradient descent method, i.e., PPO~\cite{schulman2015trust,schulman2017proximal}, to optimize the alignment process of these diffusion models. The policy gradient descent optimization process is formulated as follows:
\begin{equation}
\nabla_{\theta} \mathcal{J}'(\mathbf{\mathbf{x}}) = - \sum \frac{\nabla_\theta P_{\theta}}{P_{\theta'}}\mathcal{R}(\mathbf{x}_O) + \lambda \nabla_\theta\mathcal{KL}(P_{\theta'}|P_{\theta}),
\end{equation}
where $\mathcal{KL}(\cdot|\cdot)$ indicates the KL-divergence between two distributions. The training process is shown as Algorithm~\ref{alg1}. Note that due to the huge GPU memory and time consumption of the video diffusion process, we distribute the data generation and network training on different GPUs. Please refer to~Sec.\ref{sec:Experiments} \textbf{Settings} for more details.

\noindent \textbf{Modeling of Reward.} To measure the quality of the reconstructed event frame, we utilize the \textit{FVD} and \textit{SSIM} as the reward $\mathcal{R}(\cdot)$ to guide the training process of reinforcement alignment.
Moreover, to remove the bias of rewards, we randomly generate $M$ samples given one prompt, forming a pair of samples.

\section{Experiments}
\label{sec:Experiments}

\begin{figure}[htbp]
    \centering
    \begin{subfigure}[b]{0.333\textwidth}
        \centering
        \includegraphics[width=\textwidth]{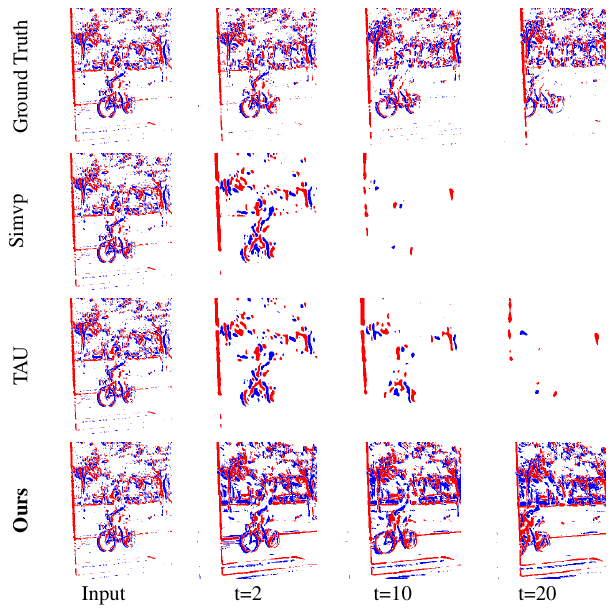}
    \end{subfigure}
    \hspace{0.001\textwidth}
    \begin{subfigure}[b]{0.320\textwidth}
        \centering
        \includegraphics[width=\textwidth]{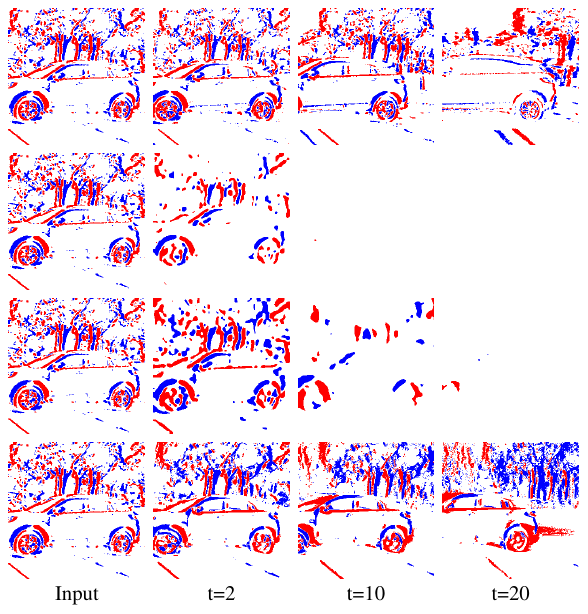}
    \end{subfigure}
    \hspace{0.001\textwidth}
    \begin{subfigure}[b]{0.320\textwidth}
        \centering
        \includegraphics[width=\textwidth]{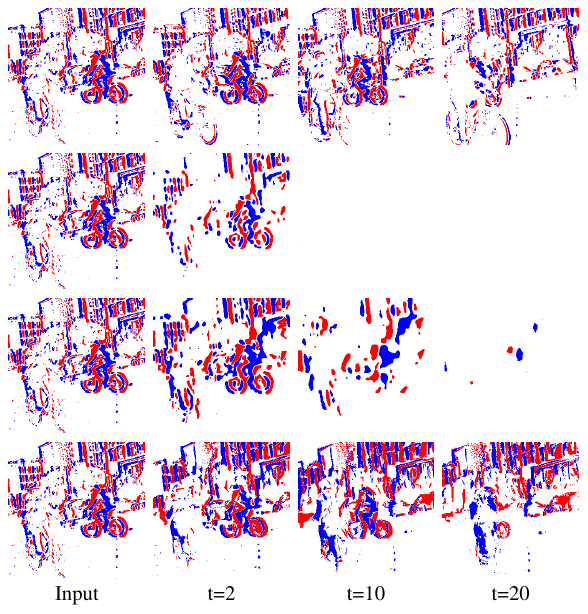}
    \end{subfigure}
    
    
    \caption{Qualitative comparison between SOTA methods. The first row of each sequence represents the ground truth of the event sequence. The second and third rows respectively depict the results of future event estimation by SimVP~\cite{gao2022simvp} and TAU~\cite{Tan_2023_CVPR}. The final row represents the results obtained by our method. The complete sequence is shown in Fig.\ref{fig:compare}.}
    \label{fig:compare_small}
    
\end{figure}

\begin{figure}[htbp]
    \centering
    \begin{subfigure}[b]{0.485\textwidth}
        \centering
        \includegraphics[width=\textwidth]{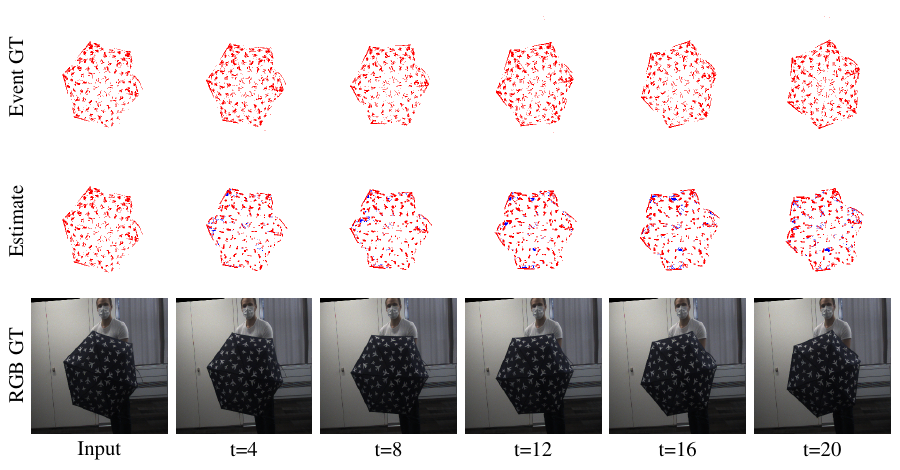}
        \caption{Rotating Umbrella.}
        \label{fig:rgb_um}
    \end{subfigure}
    \hspace{0.01\textwidth} 
    \begin{subfigure}[b]{0.485\textwidth}
        \centering
        \includegraphics[width=\textwidth]{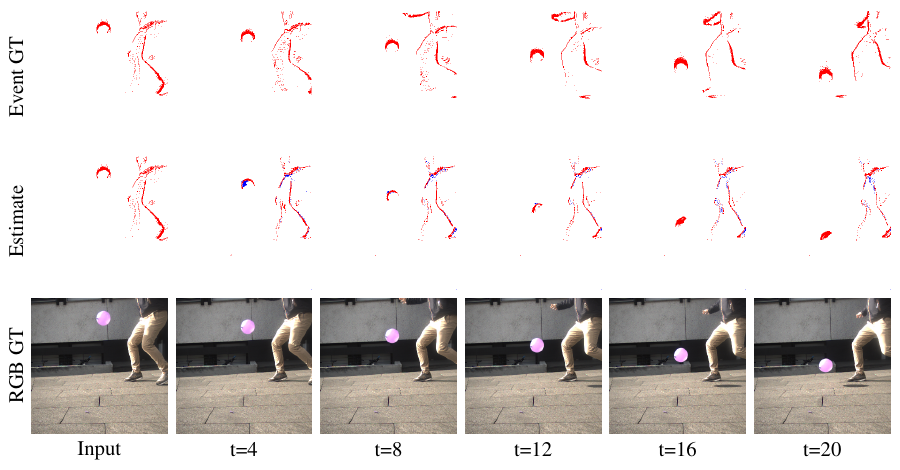}
        \caption{Falling Balloon.}
        \label{fig:rgb_ball}
    \end{subfigure}

    \vspace{0.01\textwidth}

    \begin{subfigure}[b]{0.485\textwidth}
        \centering
        \includegraphics[width=\textwidth]{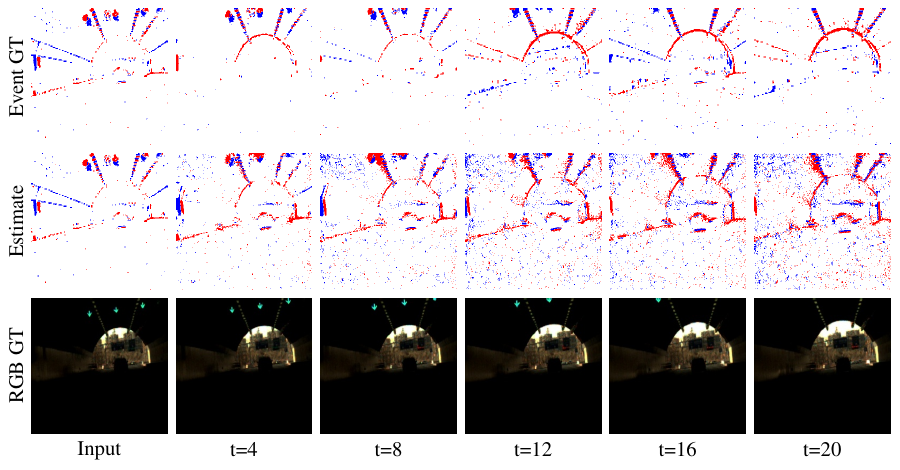}
        \caption{Poor exposure: The car driving out of the tunnel.}
        \label{fig:rgb_cave}
    \end{subfigure}
    \hspace{0.01\textwidth} 
    \begin{subfigure}[b]{0.485\textwidth}
        \centering
        \includegraphics[width=\textwidth]{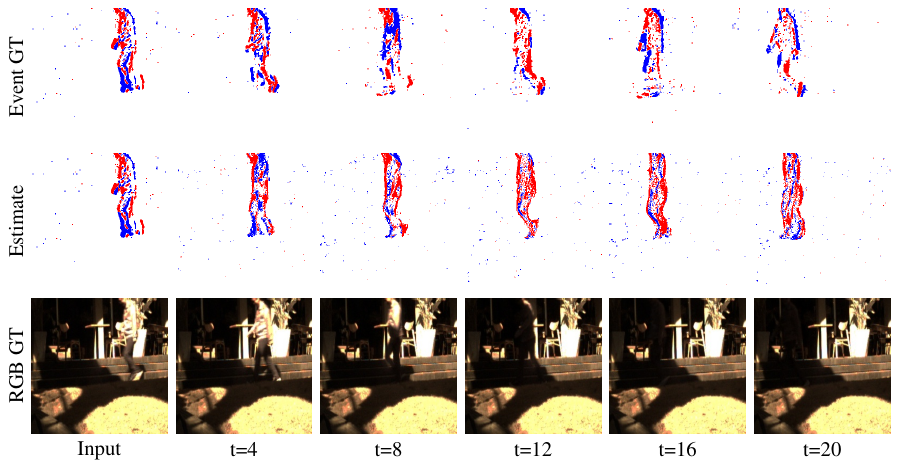}
        \caption{Occlusion: A person walking to shadow.}
        \label{fig:rgb_walk}
    \end{subfigure}
    \caption{More visualization of our method’s prediction in various scenarios. The results of the complete sequence along with other methods are presented in Fig.~\ref{fig:uzh} and Fig.~\ref{fig:various}.}
    
\end{figure}

\noindent \textbf{Dataset.} 
In our study, we utilize two large-scale event datasets, i.e., VisEvent~\cite{wang2023visevent} and EventVOT dataset~\cite{wang2023event}. The VisEvent dataset encompasses a wide variety of scenes, including 820 RGB-Event video pairs and 371,128 RGB frames. Moreover, it was captured by the DAVIS346 camera~\cite{brandli2014240}, with resolutions of $346\times260$ for RGB and events. It addresses diverse environmental conditions and includes 17 distinct attributes such as camera motion, low illumination, and background clutter, facilitating detailed performance analysis under various challenging scenarios. 

The EventVOT dataset provides 1,141 high-definition videos with 569,359 frames, making it the largest dataset in this domain. It features a resolution of $1280\times720$, encompassing 19 diverse classes of target objects. Compared to earlier datasets such as VOT-DVS, TD-DVS, and Ulster from 2016, and more recent ones like FE108 and COESOT, EventVOT offers an unprecedented scale and variety, including a substantial number of videos and a wide array of environmental conditions, aimed at improving the development and evaluation of event-based visual tracking algorithms. The dataset is meticulously annotated and divided into training (841 videos), validation (18 videos), and testing (282 videos) subsets, ensuring a comprehensive framework for robust algorithm testing and benchmarking.

\begin{figure}[htbp]
    \centering
    \begin{subfigure}[b]{1.0\textwidth}
        \centering
        \includegraphics[width=\textwidth]{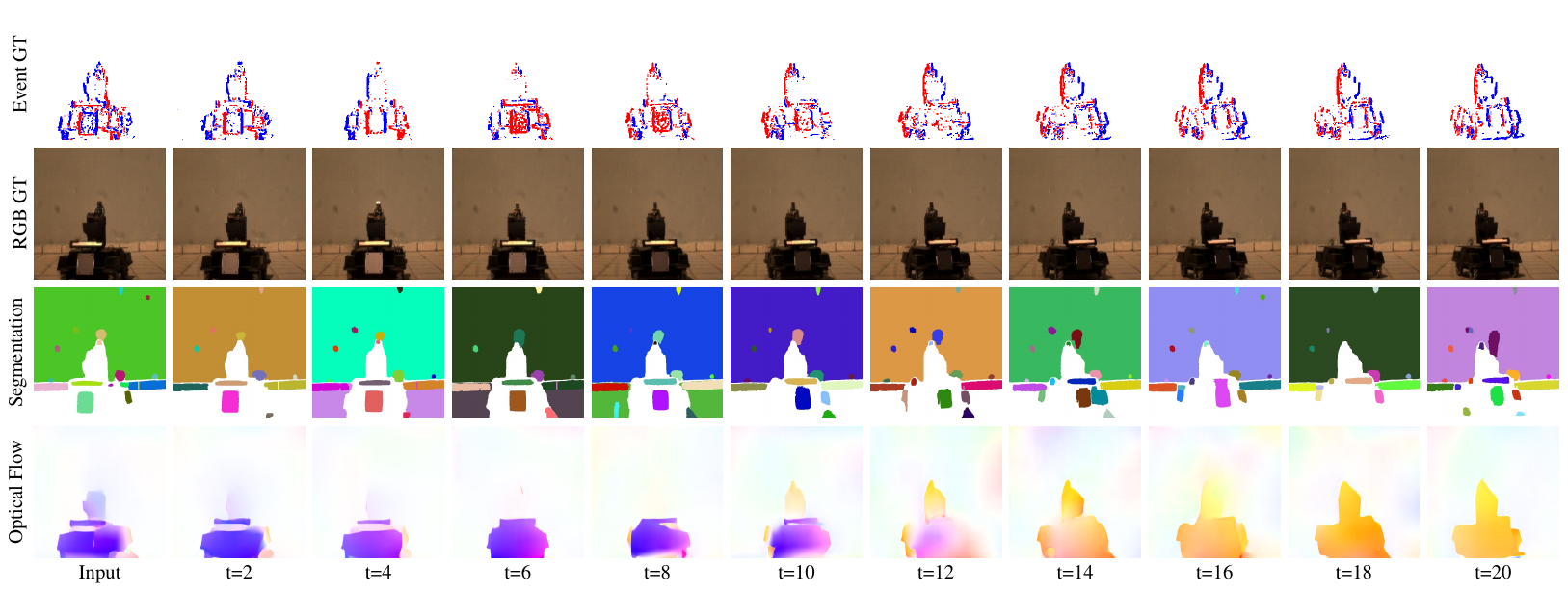}
        \caption{Segmentation and flow estimation results using GT event and RGB frame.}
        \label{fig:rgb_image1}
    \end{subfigure}

    \vspace{0.01\textwidth}

    \begin{subfigure}[b]{0.485\textwidth}
        \centering
        \includegraphics[width=\textwidth]{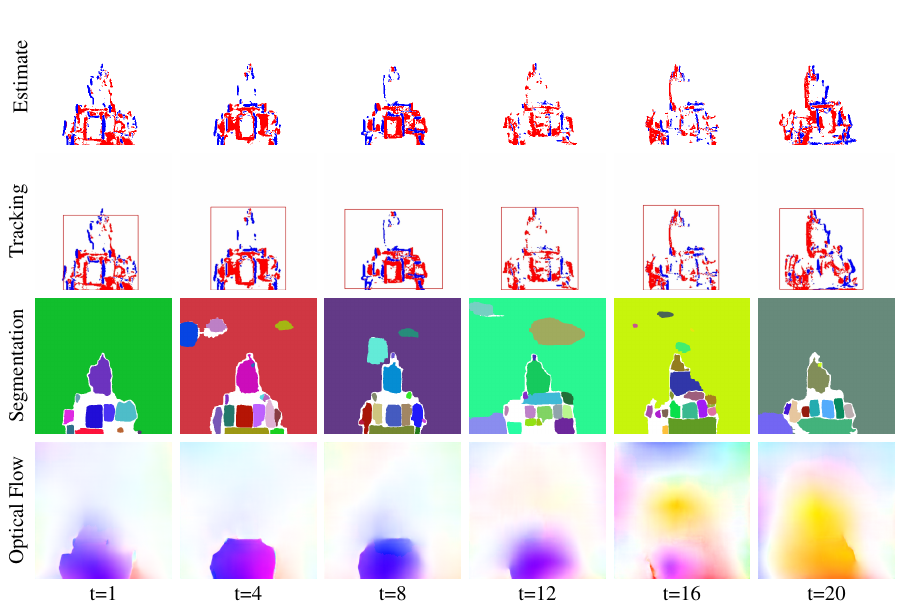}
        \caption{Downstream results using our estimated events.}
        \label{fig:rgb_image3}
    \end{subfigure}
    \hspace{0.01\textwidth} 
    \begin{subfigure}[b]{0.485\textwidth}
        \centering
        \includegraphics[width=\textwidth]{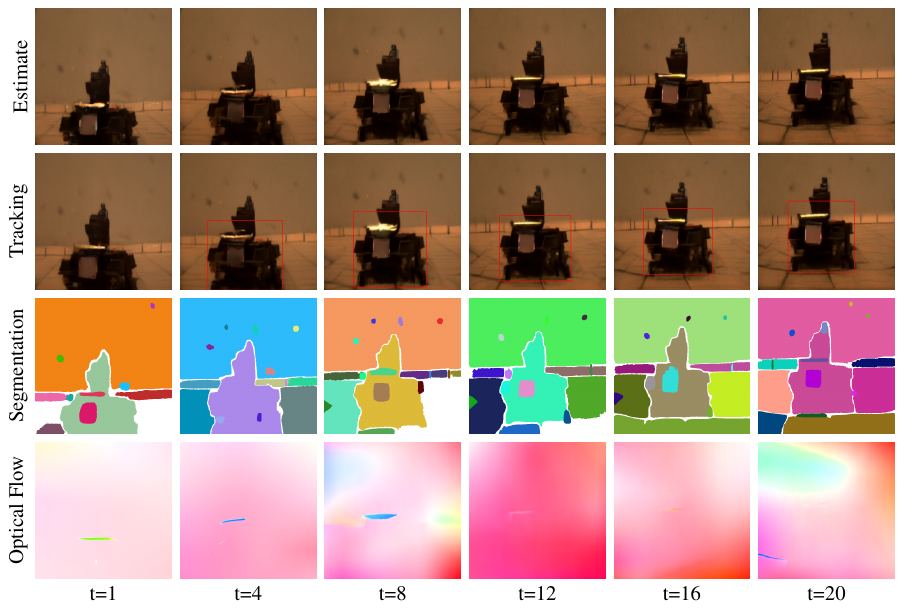}
        \caption{Downstream results using SVD estimated frames.}
        \label{fig:rgb_image4}
    \end{subfigure}
    \caption{Visualization results on downstream tasks, where we show the tasks of tracking, segmentation, and flow estimation. (a) denotes the ceiling performance of settings (b) and (c).}
    \label{fig:compare_rgb}
    
\end{figure}

\begin{table*}[t]
    \caption{Quantitative comparison between SOTA methods, where SVD denotes the standard stable video diffusion network. ``VID‘’ represents the video and ``EVT'' indicates the event data. $\uparrow$ (resp.$\downarrow$) represents the bigger (resp. lower) the better.}
    \label{tab:comparison}
    \centering
    \newcolumntype{C}{>{\centering\arraybackslash}p{1.2cm}}
    \newcolumntype{D}{>{\raggedright\arraybackslash}p{1.8cm}}
    \newcolumntype{B}{>{\centering\arraybackslash}m{0.1cm}}
    \begin{tabular}{DC|CCCCCCC}
        \toprule
        Methods & Modal & \textit{FVD}~$\downarrow$   & \textit{MSE}~$\downarrow$   & \textit{SSIM}~$\uparrow$ & \textit{LPIPS}~$\downarrow$& \textit{mIoU}~$\uparrow$ & \textit{aIoU}~$\uparrow$\\
        \midrule
        PhydNet~\cite{Guen_2020_CVPR} &VID & 1602.84 & 0.0295 & 0.4299 & 0.6048 & 0.077 & 0.348\\
        SimVP~\cite{Gao_2022_CVPR}  &VID  & 1347.57 & 0.0216 & 0.6261 & 0.3051 & 0.264 & 0.520 \\
        TAU~\cite{Tan_2023_CVPR}   &VID  & 1371.65 & 0.0240 & 0.6381 & 0.3026& 0.264 & 0.529\\
        PredRNNv2~\cite{9749915predrnn}   &VID & 1266.83 & 0.0176 & 0.5550 & 0.2407 & 0.275 & 0.540\\
        SVD~\cite{blattmann2023stable}&VID  & 1122.54 & 0.0246 & 0.6451 & 0.3299 & 0.233 & 0.506\\
        \midrule
        PredRNNv2~\cite{9749915predrnn} &EVT  & 1339.05 & 0.0306 & 0.6598 & 0.3388 &0.166 & 0.504 \\
        SimVP~\cite{Gao_2022_CVPR}  &EVT  & 1242.25 & \underline{0.0210} & 0.7961 & 0.3371 &0.213 & \textbf{0.532} \\
        TAU~\cite{Tan_2023_CVPR}    &EVT  & \underline{1218.03} & 0.0231 & \underline{}{0.7972} & \underline{0.3354} &\underline{0.228} & 0.514\\

        Ours   &EVT  & \textbf{1055.25} & \textbf{0.0170} & \textbf{0.7998} & \textbf{0.3123} &\textbf{0.302} & \underline{0.522}\\
        \bottomrule
    \end{tabular}
\end{table*}

\noindent \textbf{Settings.} All experiments are conducted on machines with $8\times$ GeForce RTX 3090 GPUs, Intel(R) Core(TM) i7-10700 CPU of 2.90GHz, and 64-GB RAM. In the pre-training stage, we employed the ADAM optimizer with the exponential decay rates $\beta_1$ = 0.9 and $\beta_2$ = 0.999. The total training process was 20000 iterations for both kinds of noise experiments. We initialized the learning rate as 1e-5. We set the batch size to 128 (with 8 gradient accumulation steps).

For the alignment process, due to the fact that it takes quite a lot of GPU memory to generate a reference trajectory during the training process, we have to achieve the trajectory generation and reinforcement alignment in a parallel and distributed manner. Specifically, we utilize $4\times$ RTX3090s to train reinforcement learning alignment processes. Moreover, $4\times$ RTX3090s is utilized to generate training trajectory data. The updating episode of the reinforcement learning process is set at 100 optimization steps. (\textit{Please refer to the Appendix for the detailed illustration of})
We also employed the ADAM optimizer with the exponential decay rates $\beta_1$ = 0.9 and $\beta_2$ = 0.999. We initialized the learning rate as $2e^{-6}$ and set the batch size to 16(with 2 gradient accumulation steps) in all experiments.

\noindent \textbf{Comparison Methods.} To comprehensively evaluate the performance of the proposed method, we compared different methods, which could be briefly divided into two categories: image-based future frame prediction and event-based future motion prediction. For the image-based future frame prediction, we adopt the original stable video diffusion~\cite{blattmann2023stable} with some popular time serious forecasting methods, e.g., PredRNN~\cite{wang2017predrnn}, SimVP~\cite{gao2022simvp}, TAU~\cite{tan2023temporal}. Moreover, for future event forecasting, we simply use stable video diffusion~\cite{blattmann2023stable} as our baseline since other methods make it hard to achieve desirable performance on those datasets.

\noindent \textbf{Metrics.} In the evaluation of our proposed model, we employ a comprehensive suite of metrics designed to assess the quality. We incorporate the \textit{Fréchet Video Distance (FVD)}~\cite{unterthiner2019fvd} to evaluate the temporal coherence and visual quality of generated video sequences against real video content. The \textit{Mean Squared Error (MSE)} serves as a fundamental metric, quantifying the average squared difference between the estimated values and the actual values. The \textit{Structural Similarity Index Measure (SSIM)} is employed to assess the visual impact of structural information, brightness, and contrast differences between the generated images and the ground truth. Lastly, the \textit{Learned Perceptual Image Patch Similarity (LPIPS)}~\cite{zhang2018unreasonable} metric is adopted to evaluate the perceptual similarity between the generated and real images. In addition, we apply \textit{mIoU} and \textit{aIoU} as metrics for the object segmentation task to validate the feasibility of the generated events in downstream tasks.

\begin{table*}[t]
    \caption{Quantitative comparison between SOTA methods on object tracking. SVD denotes the standard stable video diffusion network. ``VID‘’ represents the video and ``EVT'' indicates the event data. $\uparrow$ (resp.$\downarrow$) represents the bigger (resp. lower) the better. \textit{$R_{upper}$} denotes a comparison of task performance with GT Event, where we average the ratio of different metrics (please refer to the Appendix for the detailed comparisons).}
    \label{tab:comparison_tracking}
    \centering
    \newcolumntype{C}{>{\centering\arraybackslash}p{1.2cm}}
    \newcolumntype{D}{>{\raggedright\arraybackslash}p{1.8cm}}
    \newcolumntype{B}{>{\centering\arraybackslash}m{0.1cm}}
    \begin{tabular}{DC|CCCCC}
        \toprule
        Methods & Modal & \textit{AUC}~$\uparrow$   & \textit{OP50}~$\uparrow$   & \textit{OP75}~$\uparrow$ & \textit{PR}~$\uparrow$& \textit{$R_{upper}$}~$\uparrow$\\
        \midrule
        PredRNN~\cite{9749915predrnn} &EVT  & 28.54 & 22.24 & 17.24 & 14.96 &0.496 \\
        SimVP~\cite{Gao_2022_CVPR}  &EVT  & 35.10 & 32.41 & 18.87 & 21.74 &0.640 \\
        TAU~\cite{Tan_2023_CVPR}    &EVT  & 34.95 & 31.85 & 20.31 & 24.74 &0.669\\
        \textbf{Ours}   &EVT  & 43.77 & 47.77 & 27.31 & 30.99 &\textbf{0.891} \\
        GT Event    &EVT & 47.87 & 53.30 & 31.93 & 34.62  & 1.000\\
        \bottomrule
    \end{tabular}
\end{table*}

\subsection{Quantitative and Qualitative Comparison Results} 
The experimental outcomes are illustrated in Table~\ref{tab:comparison}. It is evident that by applying regularization to event data, our methodology attains performance levels comparable to state-of-the-art (SOTA) methods. Notably, in metrics specific to the event domain, our approach outperforms the comparative method of SVD. The visual results, as shown in Fig.~\ref{fig:compare_small}, demonstrate that compared to TAU, SimVP, and other methods, our approach can predict longer durations, more precise motion, and generate events that are more stable and closer to the ground truth.

In addition to the standard tests, we further validated our method on the BS-ERGB~\cite{Tulyakov21cvpr} dataset, which was captured using a completely different event camera. As shown in Fig.~\ref{fig:rgb_um} and Fig.~\ref{fig:rgb_ball}, our method achieves satisfactory prediction results on this dataset, demonstrating its effectiveness. Furthermore, we conducted experiments in extreme scenarios, with the visualization results shown in Fig.~\ref{fig:rgb_cave} and Fig.~\ref{fig:rgb_walk}. Even under stringent lighting conditions, our method is still able to predict human and viewpoint motion with high accuracy, indicating strong robustness.

For downstream task results, we conducted both qualitative and quantitative evaluations on the tasks of target segmentation and object tracking, and performed qualitative evaluations on optical flow estimation. Fig.~\ref{fig:compare_rgb} shows the qualitative results on downstream tasks. Due to the motion information priors provided by the \textbf{high temporal resolution event prompt}, our method can accurately predict the tank's turning direction, whereas using only a single input RGB frame makes this difficult.  Since directly comparing cross-modal results does not provide meaningful insights, we defined an upper bound ratio $R_{upper}$ based on the upper bound of target tracking performance for comparison. Table \ref{tab:comparison_tracking} presents the quantitative comparison results, showing that our method significantly outperforms other methods. \textit{For more detailed definitions and results, please refer to the Appendix materials.}

\begin{table*}[t]
    \newcolumntype{C}{>{\centering\arraybackslash}p{1.2cm}}
    \caption{Ablation Study of Pre-training Phase. All models are tested with only feeding single event voxel frame, where ''\#Prompt'' indicates the number of training prompt event data, $\mathcal{U}(1,3)$ indicates to randomly select 1 to 3 frames as training prompt. 'Fine-tuning' refers to the parameters that are fine-tuned, 'T' indicates the fine-tuning of all temporal attention parameters, and 'S+T' indicates the simultaneous fine-tuning of both spatial and temporal attention parameters. 'CLIP' indicates whether the CLIP features extracted have been fine-tuned for events. ’RGB’ refers to the CLIP model pre-trained on RGB data, while ‘Event‘ indicates the CLIP model fine-tuned with event data.}
    \label{Tab:Train-config}
    \centering
    \begin{tabular}{cccc|CCCCC}
        \toprule
        &\#Prompt & Fine-tuning & CLIP & \textit{FVD}~$\downarrow$    & \textit{SSIM}~$\uparrow$ & \textit{LPIPS}~$\downarrow$  & \textit{mIoU}~$\uparrow$ & \textit{aIoU}~$\uparrow$\\
        \midrule
        &$\mathcal{U}(1,3)$  & T & RGB  & 1972.91 & 0.69513 & 0.3651 & 0.266 & 0.507 \\
        &$\mathcal{U}(1,3)$  & S+T & RGB & \textbf{1378.92} & \textbf{0.78496} & \textbf{0.3076}& 0.252 & \textbf{0.525} \\
        &$1$  & S+T & RGB & 1406.24 & 0.78374
        & 0.3219   & \textbf{0.268} & 0.524 \\
        &$1$  & S+T & Event & 1646.78 & 0.72779 & 0.3464   & - & - \\
        \bottomrule
    \end{tabular}
\end{table*}

\begin{table*}[t]
    \newcolumntype{C}{>{\centering\arraybackslash}p{1.2cm}}
    \caption{Ablation Study of motion alignment and multi prompt. All models are tested with only feeding single event voxel frame. 'EP' denotes denoising using the high temporal resolution event prompt, and 'MA' denotes motion alignment based on reinforcement learning.}
    \label{Tab:ablation}
    \centering
    \begin{tabular}{ccccCCCCC}
        \toprule
        & Method &\textbf{EP} &\textbf{MA} & \textit{FVD}~$\downarrow$  & \textit{SSIM}~$\uparrow$ & \textit{LPIPS}~$\downarrow$ & \textit{mIoU}~$\uparrow$ & \textit{aIoU}~$\uparrow$ \\
        \midrule
        & A  & $\times$ & $\times$ & 1378.92  & 0.78496 & \textbf{0.3076} & 0.252 & 0.505\\
        & B  & $\checkmark$ & $\times$ & 1227.56  & 0.79077 & \underline{0.3101} & \underline{0.295} & \textbf{0.528} \\
        & C  & $\times$ & $\checkmark$ & \underline{1119.71} &\underline{0.79597}  & 0.3246  & 0.277 & 0.516  \\
        & D  &$\checkmark$ & $\checkmark$& \textbf{1055.25} & \textbf{0.79981} & 0.3123 &\textbf{0.302} & \underline{0.522} \\
        \bottomrule
    \end{tabular}
\end{table*}

\subsection{Ablation Study}
\noindent \textbf{Fine-tuning Layers.} 
As delineated in Table~\ref{Tab:Train-config}, we conduct empirical validations across various fine-tuning configurations, including temporal-only and joint spatial-temporal adjustments. Note that, during the refinement process, we keep the neural network with a similar amount of training weights. Moreover, all methods are trained under the same configuration, excluding the training parameters. The results from these experiments provide critical insights into the optimal configuration settings that enhance the accuracy and efficiency of event-based video generation. 
Besides, the authors also want to note that it's ineffective to only change some parts of a large generative model, since the diffusion U-Net is trained with the perception of original VAE, CLIP models generation distributions. We also have experimentally validated that after changing those modules. The experimental results are shown in Table~\ref{Tab:Train-config}, where we feed features from different clip models (Event-trained or RGB-trained) to the SVD U-Net. Note that all CLIPs are fed with event voxels. Even further fine-tuning the SVD with plenty of data, the resulting diffusion model with Event-trained is still underperformed.

\noindent \textbf{Testing-time Prompt Augmentation.} As detailed in Sec.~\ref{Sec:Pretrain}, we enhance the test-time prompt by incorporating multiple event frames of high temporal resolution. To thoroughly examine the impact of these hyperparameters and ascertain the efficacy of our test-time augmentation strategy, we compare various approaches employing different test-time prompts, as delineated in Table~\ref{table:Testing-time}, titled 'Testing-time'. From this comparative analysis, it is evident that there is a gradual improvement in the neural network's performance as the extent of prompt augmentation increases.

\subsection{Discussion}
While the proposed method has exhibited preliminary capabilities, it is imperative to address its existing limitations. One significant challenge is the inherent nature of event data; while it boasts high temporal resolution, this type of data typically lacks texture, thereby impeding the effective capture of detailed semantic information. This limitation highlights the difficulty in accurately representing complex visual scenes solely based on event data. Furthermore, there is a pressing need for the development of a lossless representation technique that can fully preserve the unique high-temporal attributes of event data, ensuring no critical information is lost during processing. \textit{Specific constrained scenarios are visualized and discussed in detail in Appendix Section \ref{limit}.}

To address these challenges and enhance the efficacy of the proposed method, future work should focus on several key areas. Firstly, advancing the method's ability to interpret texture would mark a significant improvement. This could involve integrating additional sensory inputs, e.g., RGB images, or employing more sophisticated data fusion techniques. Secondly, \textbf{the creation of a more comprehensive event sequence dataset is essential}. Such a dataset should encompass a wider variety of scenarios and conditions, thereby providing a robust platform for training and testing the improved models. By addressing these aspects, future research can pave the way for more accurate, reliable, and versatile event-based vision systems for object motion forcasting.

\section{Conclusion}
\label{Sec:conclusion}
In this study, we have introduced the Event-Sequence Diffusion Network, a novel approach poised to redefine the landscape of video diffusion. By leveraging event-based data, characterized by its high temporal resolution, our methodology advances the frontiers of motion modeling, enabling the generation of video sequences that are not only rich in detail but also grounded in the realistic dynamics of object motion. Our approach stands in stark contrast to traditional video diffusion models that rely on single images or textual prompts for initialization. By employing sequences of events as the conditioning input, we ensure a more nuanced and temporally coherent synthesis of video content.

Future work will focus on refining the Event-Sequence Diffusion Network, exploring its applicability across a broader spectrum of computer vision tasks, and enhancing its efficiency for real-time applications. Moreover, we aim to delve deeper into the interplay between event-based data and diffusion models, seeking to unlock new potentials and applications in areas such as autonomous navigation, interactive gaming, and dynamic scene reconstruction.

\clearpage

\medskip

\bibliography{main}

\newpage

\appendix

\section*{Appendix Overview}

In this Appendix material, we provide additional details to complement the content of the paper, including the training and inference details of our method$\textbf{(Sec.\ref{sec:details})}$, an analysis of the results from comparative methods, more experiments on downstream tasks$\textbf{(Sec.\ref{sec:down})}$, details related to reinforcement learning$\textbf{(Sec.\ref{sec:sup_rein})}$, and limitation of proposed method$\textbf{(Sec.\ref{limit})}$.


\section{Training and Inference Details}
\label{sec:details}
We provide a comprehensive training pipeline along with detailed parameter specifications, encompassing the model architecture and fine-tuning parameters.
For voxelized event data, we perform preprocessing and data augmentation to enhance the performance and generalization capability of the model.
\subsection{Architecture}
\begin{figure}[ht]
    \centering
    \resizebox{1.0\textwidth}{!}{\includegraphics{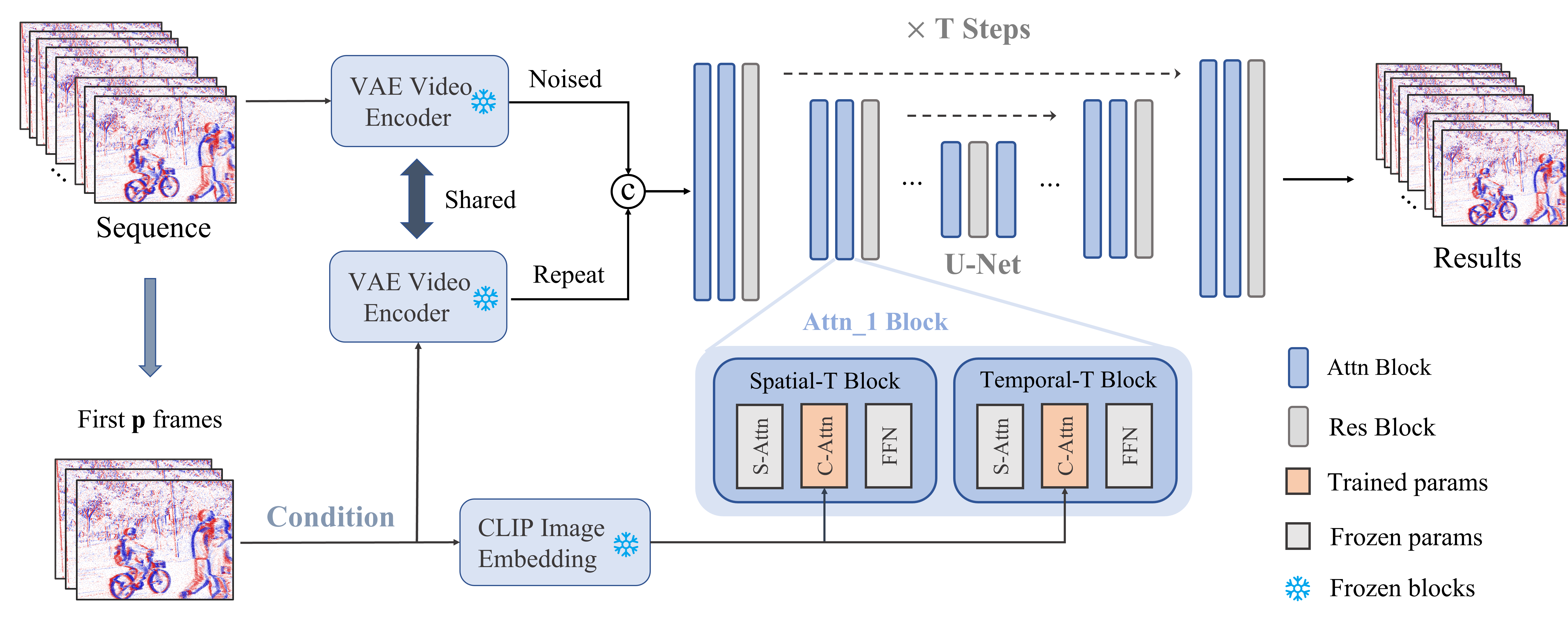}}
    \caption{Training workflow of the proposed method, where the left upper one indicates the target noised latent and the left lower one represents the prompted event sequence. We concatenate the prompt information with denoising latent for noise learning. Moreover, the feature from the CLIP model is also injected into the diffusion U-Net.}
    \label{Fig:arch}
    \vspace{-0.2cm}
\end{figure}

As discussed in Section.~\ref{Sec:Pretrain},  We employ the temporal U-Net architecture based on \textit{Stable Video Diffusion}. During training, we preprocess the event sequence into multiple frames of voxel data, which serve as conditional inputs to the network. Fig.~\ref{Fig:arch} illustrates the overall training procedure, which bears resemblance to SVD. The initial segment of the event sequence serves as the conditional input to the U-Net model. For the parameters trained in the U-Net, we only fine-tuned the cross-attention layer that incorporates the event condition. We conducted ablation experiments on different attention layers, as shown in Table~\ref{Tab:Train-config}. Simultaneously fine-tuning both spatial and temporal attention layers yielded the best results.

\subsection{Datasets and augmentation}

During the pre-training phase of the proposed diffusion model, events of poor quality can adversely affect model training. Therefore, we emulate the approach taken in EventSAM~\cite{chen2023segment} by filtering out sequences with severe degradation in VisEvent. Finally, we integrated 291 event sequences from the VisEvent dataset and 841 sequences from the EventVOT dataset. We omitted specific sequences from the CeleX-V sensors due to their absence of polarity information. For the alignment phase, we uniformly sourced all data, the same as the training dataset, aligning them in accordance with the pre-training prompts. In the evaluation stage, a random selection of 100 sequences from each dataset was employed to rigorously assess the efficacy of our proposed methodology.

During training, we randomly crop the Visevent sequences using a $256\times256$ kernel and then resize them to $128\times128$ before feeding them into the network. For the higher-resolution EventVOT dataset, in order to preserve as much motion information as possible, we randomly select crop kernels ranging from 512 to 720, crop the images accordingly, and then resize them to $128\times128$ as well. We divide the events into intervals of 20ms.

 After events are converted into voxels, normalization is required. Simple normalization in the range of 0 to 1 may lead to flickering in the sequence. Therefore, when obtaining the sequence, we perform a smoother normalization using maximum value normalization. Specifically, dividing each frame of the sequence by the maximum absolute value of the sequence ensures that the voxel values are between 0 and 1 and remain stable.

\section{Downstream tasks}
\label{sec:down}
To validate the reliability of the generated event voxels, we assessed our approach's performance on several classic downstream tasks such as object tracking, segmentation, and optical flow estimation.
\subsection{Evaluation Methods and Metrics}

For the object tracking task, we utilized the OSTrack~\cite{ye2022ostrack} library to evaluate the event voxels estimated by our method and the RGB frames estimated by SVD~\cite{blattmann2023stable}, focusing on the selected VisEvent test set to ensure alignment. Due to the significant modality differences between event data and RGB data, directly comparing tracking metrics would be unfair. Therefore, we tested on the ground truth of the selected test sequences separately for each modality, using these as performance upper bounds. The model performance was assessed by analyzing the discrepancies between the obtained results and these upper bounds. We utilized AUC (Area Under the Curve),OP50, OP75 and PR(Precision and Recall) as our base evaluation metrics,upon which we calculated the mean ratio of each metric to its upper bound, obtaining a cross-modality tracking evaluation metric $R_{upper}$:

\begin{equation}
R_{upper} = Mean(\frac{AUC}{AUC_{upper}}+\frac{OP50}{OP50_{upper}}+\frac{OP75}{OP75_{upper}}+\frac{PR}{PR_{upper}})    
\end{equation}

For the object segmentation task, we employed SAM~\cite{kirillov2023segany} to segment objects in the RGB domain and fine-tuned EventSAM~\cite{chen2023segment} on event sequences for object segmentation in the event domain. Given the absence of segmentation ground truth in the selected dataset and the strong generalization capabilities of the segmentation models in their respective domains, we used the sequence ground truth segmentation results as the ground truth for our evaluations. we employ mIOU (mean Intersection over Union) and aIOU (average Intersection over Union) as evaluation metrics.

Optical flow estimation is particularly challenging due to the limited and narrowly distributed datasets with optical flow ground truth. Although event-based optical flow estimation has been explored, it often suffers from poor generalization performance. Consequently, we restricted our qualitative experiments to simple motion scenarios involving a single object to better understand the model's performance under controlled conditions. The DDVM~\cite{saxena2024surprising} was employed to estimate the optical flow for both event and RGB sequences.

\subsection{Results Analysis}
The quantitative results of the tracking are presented in Table~\ref{tab:comparison_tracking}. Due to the modality differences, the overall tracking performance of RGB surpasses that of event tracking, as evident from their respective ground truth sequences. However, our method exhibits a higher ratio of upper bounds compared to SVD, indicating that our estimated results are closer to the ground truth, making the motion estimation more reliable.

\begin{figure}[htbp]
    \centering
    \begin{subfigure}[b]{0.485\textwidth}
        \centering
        \includegraphics[width=\textwidth]{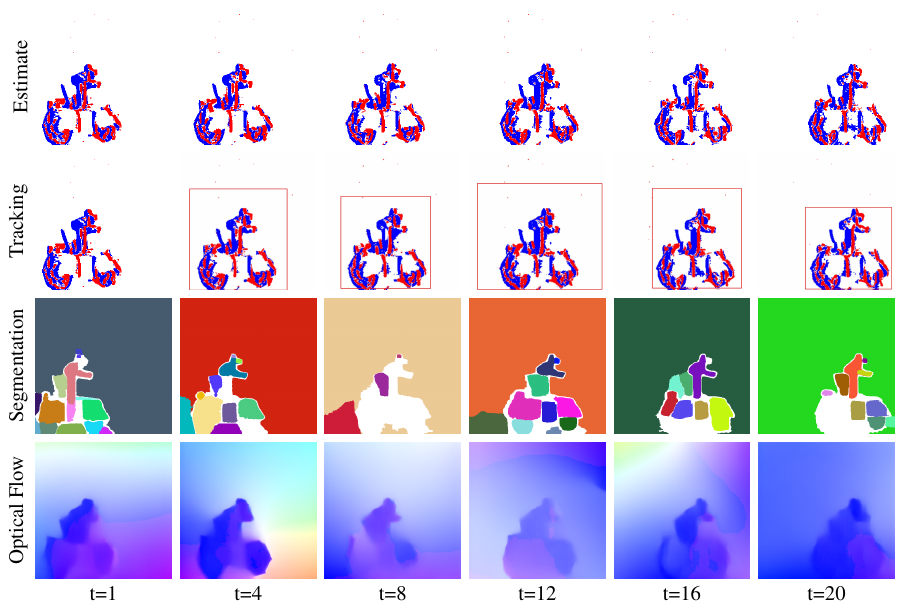}
        \caption{A tank moving towards the right.}
        \label{fig:image1}
    \end{subfigure}
    \hspace{0.01\textwidth}  
    \begin{subfigure}[b]{0.485\textwidth}
        \centering
        \includegraphics[width=\textwidth]{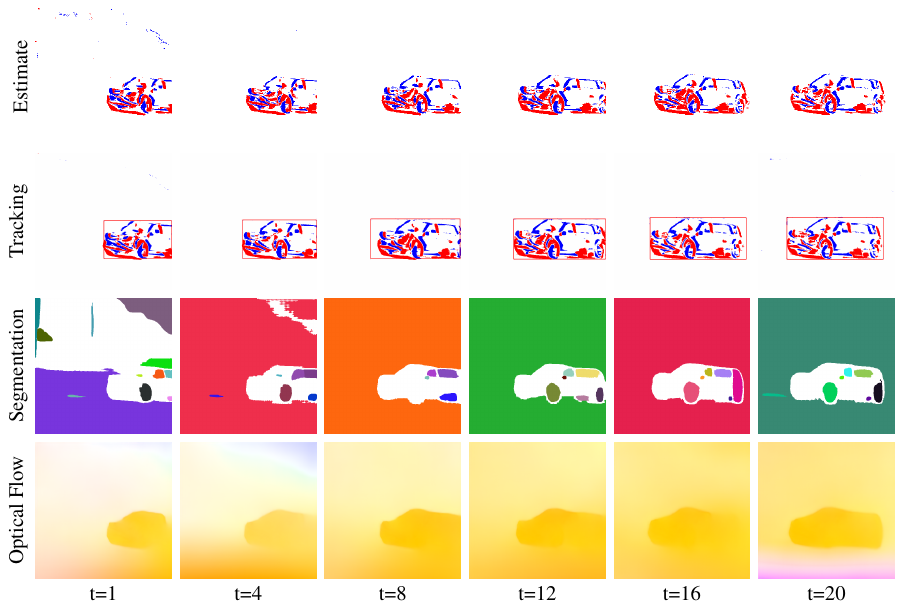}
        \caption{A car moving towards the left.}
        \label{fig:image2}
    \end{subfigure}
    
    \caption{More results of downstream tasks.}
    \label{fig:more_res}
\end{figure}

Fig.~\ref{fig:compare_rgb} and Fig.~\ref{fig:more_res} depict the visual results of object tracking. In Fig.~\ref{fig:rgb_image4}, due to erroneous motion trajectory prediction by SVD, a significant discrepancy between tracking and labeling is observed. Conversely, our method demonstrates better trajectory estimation, attributed to the high temporal resolution of event prompts providing prior information on motion during denoising.

Table~\ref{tab:comparison} presents the quantitative results of object segmentation, where our method achieves the best performance. Additional visual results are provided in Fig.~\ref{fig:segmentation}, illustrating that the event segmentation generated by our method closely resembles the ground truth.

Optical flow is typically employed to describe object motion. In our work, the emphasis of optical flow estimation lies in the accuracy of motion prediction. Fig.~\ref{fig:compare_rgb} and Fig.~\ref{fig:more_res} illustrate that our method can predict motion more precisely, leading to better optical flow estimation results.

\section{Reinforcement learning for SVD}
\label{sec:sup_rein}
\subsection{Event voxel normalization based on standard deviation.}
In reinforcement learning, the choice of the reward function is a crucial factor in determining training outcomes. We selected the FVD and SSIM metrics, which are closest to perceptual results, as our reward functions.Specifically, the final reward function is defined as $R(x)=SSIM(x)+\lambda(FVD(x))$, where $\lambda$ equals 2. Upon each model update, reward scores of the data generated from the previous batch are normalized to a standard normal distribution.

Further, To quantitatively demonstrate the rationale behind our selection of these two metrics, we actually have experimentally tried to utilize a mixture of all metrics to model the reconstruction reward. However, the performance results are much worse than the method we ultimately used, as shown in Table.~\ref{table:select}. It may be due to that sometimes the optimization of pixel-level metrics of MSE and PSNR may be contradictory to perceptual metrics. Thus, such a mixture of metrics makes the objective hard to optimize. Meanwhile, due to the target data distribution, it's more plausible to optimize the process on the data manifold (perceptual space) than the raw space.

\begin{table*}[ht]
\centering
\caption{Ablation Study of reward metrics}
\label{table:select}
\begin{tabular}{lcccccc}
\toprule
\textbf{Reward Metrics} & \textit{MSE}~$\downarrow$ & \textit{PSNR}~$\uparrow$   & \textit{FVD}~$\downarrow$ & \textit{FID}~$\downarrow$ & \textit{SSIM}~$\uparrow$ & \textit{LPIPS}~$\downarrow$ \\
\midrule
mixture metrics & 0.0240 & 16.198 & 1562.66 & 265.32 & 0.6674 & 0.3463 \\
FVD \& SSIM    & 0.0170 & 17.696 & 1055.25 & 243.45 & 0.7998 & 0.3123 \\
\bottomrule
\end{tabular}
\end{table*}

Directly utilizing spatial metrics as evaluation criteria in the RGB domain is typically feasible because RGB data contains dense spatial information, and the distribution of motion and static scenes is relatively similar. However, this approach is not viable for the event domain, which only records motion information. In Fig.~\ref{fig5:image2}, the samples illustrate this point. As depicted in the rightmost column, when capturing a static scene, the event camera does not generate events, leading to a high original reward score. However, the samples in the first two columns occur in scenarios with significant motion, resulting in relatively high-quality, natural outcomes. Nevertheless, when normalized alongside the results from the first column, even these samples yield negative scores.

Fig.~\ref{fig:curve} illustrates the distribution of reward scores with respect to sample standard deviation. Typically, higher standard deviations in samples correspond to lower scores, indicating poorer performance. Conversely, in relatively static scenes, higher scores are typically obtained. Such outcomes may inadvertently induce an overall trend toward static motion, which is not desirable. As illustrated by the green curve in Fig.~\ref{fig:curve}, after several model parameter updates, the reward score begins to decline, ultimately resulting in unsatisfactory outcomes.

To address this issue, we perform normalization of samples based on their standard deviation each time parameters are updated. Specifically, the standardized $SCORE_{std}$ denoted as $SCORE_{std}=SCORE+\beta\left(std(x)-std_{min}\right)$, where $std_{min}$ represents the minimum value of the total standard deviation of the previous batch of generated samples and $\beta$ equals 30.

\subsection{Training Setting and Results Analysis}

\begin{figure}[htbp]
    \centering
    \begin{subfigure}[b]{0.29\textwidth}
        \centering
        \includegraphics[width=\textwidth]{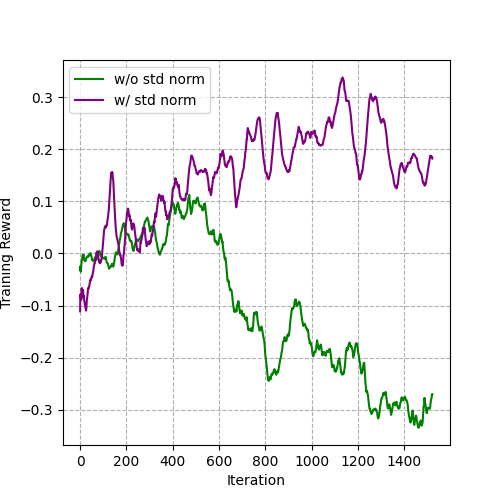}
        \caption{}
        \label{fig:curve}
    \end{subfigure}
    \hspace{0.01\textwidth}  
    \begin{subfigure}[b]{0.35\textwidth}
        \centering
        \includegraphics[width=\textwidth]{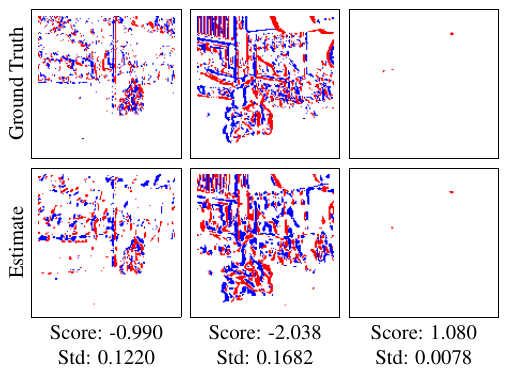}
        \caption{}
        \label{fig5:image2}
    \end{subfigure}
    \hspace{0.01\textwidth}  
    \begin{subfigure}[b]{0.31\textwidth}
        \centering
        \includegraphics[width=\textwidth]{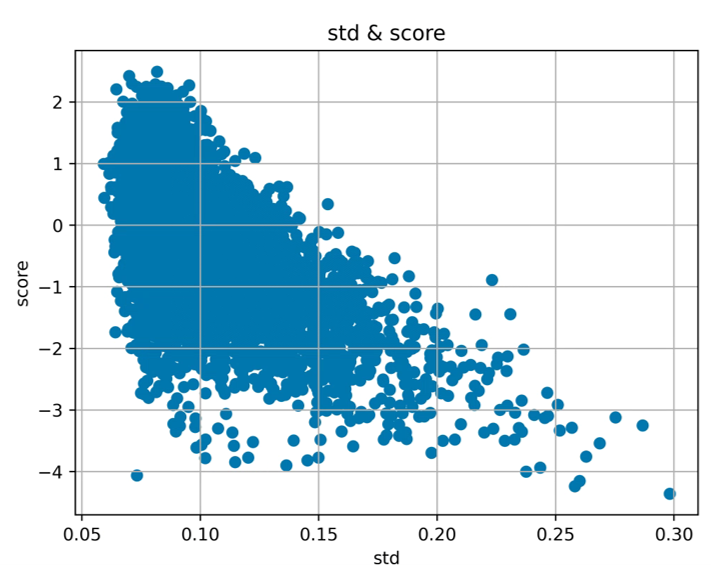}
        \caption{}
        \label{fig5:image3}
    \end{subfigure}

    \caption{Analysis of the Reinforcement Learning Process. Fig.~\ref{fig:curve} illustrates the reward curves during training, with the purple curve representing the scenario with standard deviation normalization applied, and the green curve representing the scenario without it.Fig.~\ref{fig5:image2} displays partial visualization results during the training process, where the first row represents the ground truth, and the second row depicts the results estimated by the pre-trained model. Fig.~\ref{fig5:image3} illustrates the distribution of reward scores with respect to standard deviation normalization for all training samples.}
    \label{fig:reinforce}
    
\end{figure}

During the training process, we initialize the training with the pre-trained model obtained from Section~\ref{Sec:Pretrain}. The batch size for training is set to 64, and the model is updated every 100 iterations, simultaneously updating the sample pool. Gaussian normalization and standard deviation normalization are sequentially applied to the sample pool. After standardizing the samples, the training process returns to a positive trajectory. The purple curve in Fig.~\ref{fig:curve} represents the score curve during training, with convergence observed around the 1000th iteration.

Fig.~\ref{fig:rein_vis} illustrates the visual results before and after reinforcement learning. Due to the inherent unpredictability and randomness in the generation of results by diffusion, utilizing only the results generated by the pre-trained model (Fig.~\ref{fig6:image2}) often leads to uncontrollable distortions and deformations. However, the results after reinforcement learning (Fig.~\ref{fig6:image3}) tend to be more stable, with motion trends closely resembling real-world scenarios.

\begin{algorithm}[t]
  \caption{Multi Prompt Reverse Process}
  \label{alg2}
  \begin{algorithmic}[1]
        \vspace{1.2mm}
        \State $\mathbf{X}_T^{pm} \sim \mathcal{N}(0, \mathbf{I})$,$\mathbf{X}_T^{rc} \sim \mathcal{N}(0, \mathbf{I})$ 
        \State \textbf{For} $t=T,...,1$ \textbf{do}
        \State ~~~~\textbf{if} $t\geq \tau$ \textbf{then}
        \State ~~~~$\mathbf{X}_{t-1}^{pm} \leftarrow  \sqrt{\Bar{\alpha}_t} \mathbf{X}_0^{pm} + \sqrt{1 - \Bar{\alpha}_t}\mathbf{X}_T^{pm}$.
        \State ~~~~$\mathbf{X}_{t}^{rep} \leftarrow  \textbf{replace}\left(\mathbf{X}_{t}^{pm},\mathbf{X}_{t}^{rc}\right)$.
        \State ~~~~Sampling $\mathbf{X}_{t-1}^{rc}$ from $\mathbf{X}_{t}^{rep}$ via Eq.~\ref{eq:DDIM}.
        \State ~~~~\textbf{else} Sampling $\mathbf{X}_{t-1}^{rc}$ from $\mathbf{X}_t^{rc}$ via Eq.~\ref{eq:DDIM}.
        \State \textbf{End for}
        \State \textbf{Return} $\mathbf{X}_0$
        \vspace{1.2mm}
  \end{algorithmic}
\end{algorithm}

\begin{table*}[t]
    \newcolumntype{C}{>{\centering\arraybackslash}p{1.4cm}}
    \caption{Ablation study of different number of testing-time prompts. Here we only compare with the nearest 10 future event voxels. All experimental settings share the same model weights. }
    \label{table:Testing-time}
    \centering
    \begin{tabular}{cc|CCCCCC}
        \toprule
        &Metric  & 1 & 2 & 4 & 8 & 12 & 15 \\
        \midrule
        &\textit{MSE}~$\downarrow$        & 0.0177 & 0.0195 & 0.0170 & 0.0196 & 0.0178 & 0.0182 \\
        &\textit{FVD}~$\downarrow$        & 1170.48 & 1196.43 & 1055.25 & 1137.69
        & 1142.15 & 1188.01\\
        &\textit{FID}~$\downarrow$        & 242.318 & 247.803 & 243.451 & 243.079 & 238.369 & 239.297 \\
        &\textit{SSIM}~$\uparrow$         & 0.79349 & 0.78988 & 0.79981 & 0.80718  & 0.82553 & 0.83545 \\
        &\textit{LPIPS}~$\downarrow$      & 0.3272 & 0.3228 & 0.3123 & 0.3228 
        & 0.3189 & 0.3308 \\
        &\textit{mIoU}~$\uparrow$      & 0.287 & 0.296 & 0.302 & 0.302 
        & 0.309 & 0.302 \\
        &\textit{aIoU}~$\uparrow$      & 0.518 & 0.527 & 0.522 & 0.529 
        & 0.525 & 0.518 \\
        \bottomrule
    \end{tabular}
\end{table*}

\begin{figure}[htbp]
    \centering
    \begin{subfigure}[b]{1.0\textwidth}
        \centering
        \includegraphics[width=\textwidth]{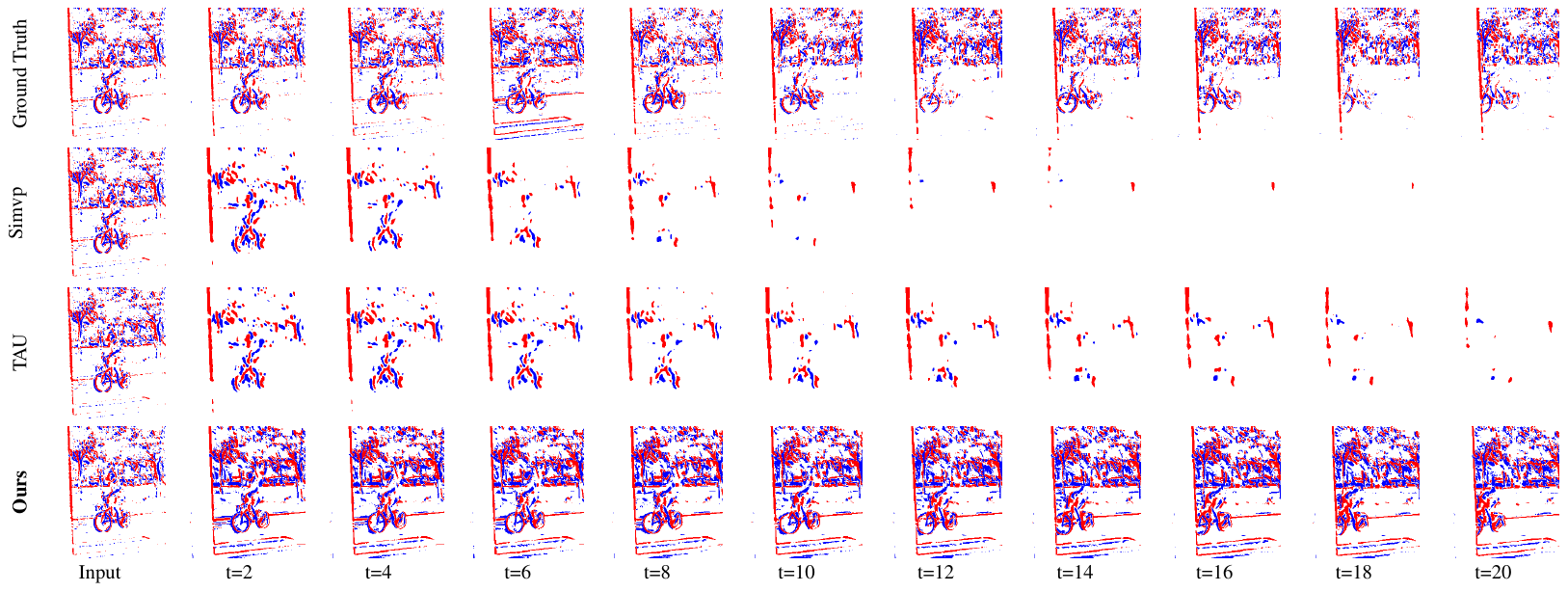}
        \caption{A bicycle riding into an obstacle.}
    \end{subfigure}

    \vspace{0.01\textwidth}

    \begin{subfigure}[b]{1.0\textwidth}
        \centering
        \includegraphics[width=\textwidth]{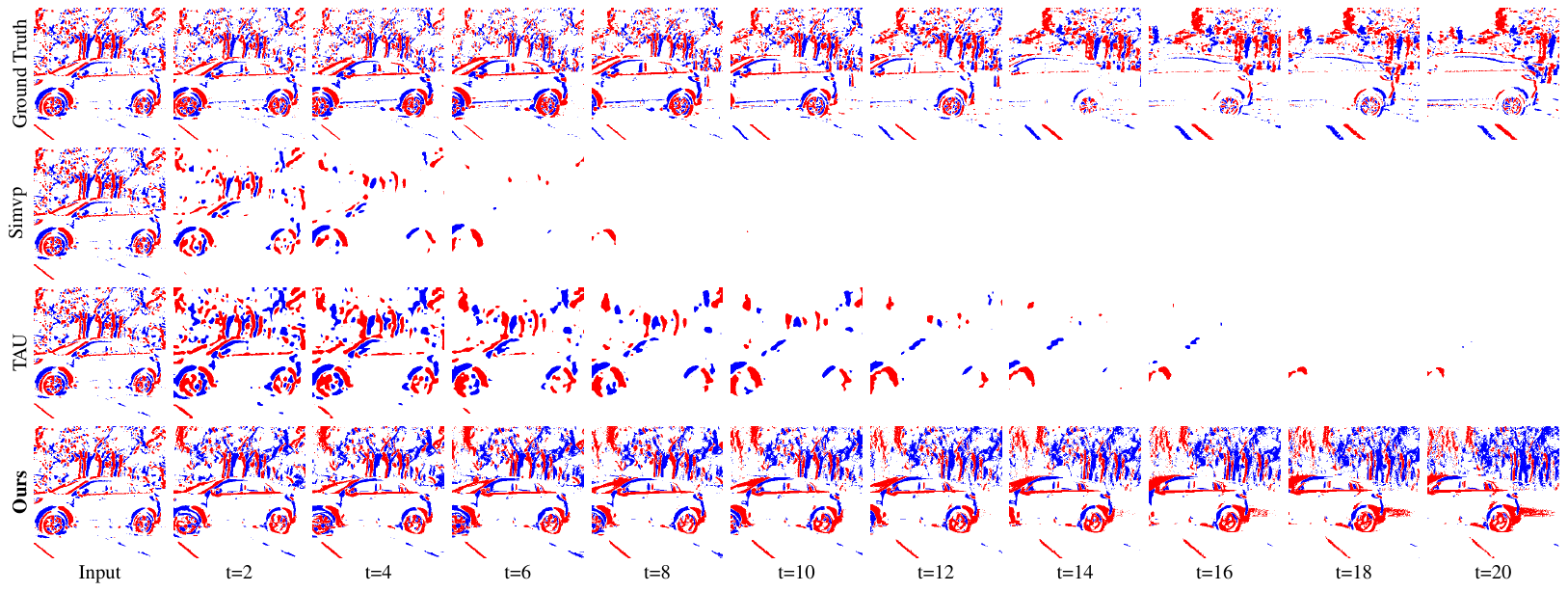}
        \caption{A car driving to the left, with the camera lens rotating to follow.}
    \end{subfigure}

    \vspace{0.01\textwidth}
    
    \begin{subfigure}[b]{1.0\textwidth}
        \centering
        \includegraphics[width=\textwidth]{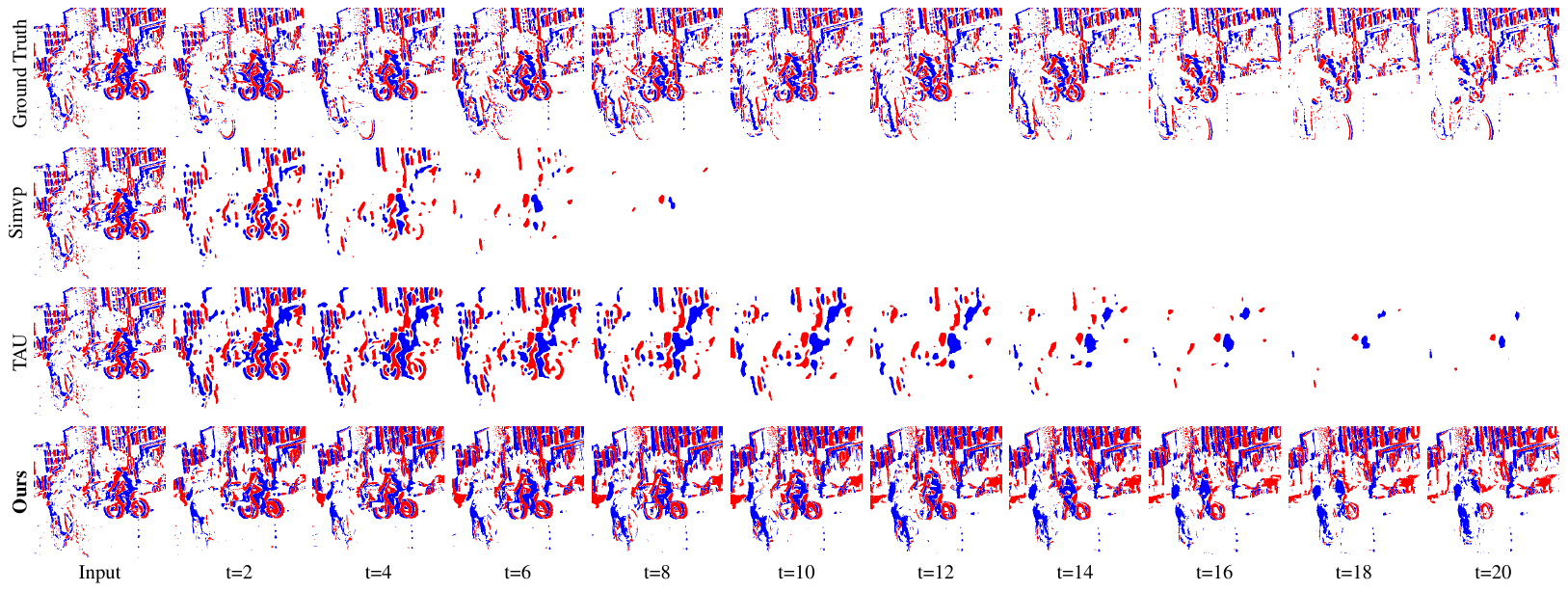}
        \caption{Two bicycles traveling towards each other and intersecting.}
        \label{fig4:image3}
    \end{subfigure}

    \caption{Qualitative comparison between SOTA methods. The first row of each sequence represents the ground truth of the event sequence. The second and third rows respectively depict the results of future event estimation by SimVP~\cite{gao2022simvp} and TAU~\cite{Tan_2023_CVPR}. The final row represents the results obtained by our method.}
    \label{fig:compare}
    
\end{figure}

\begin{figure}[htbp]
    \centering
    \begin{subfigure}[b]{1.0\textwidth}
        \centering
        \includegraphics[width=\textwidth]{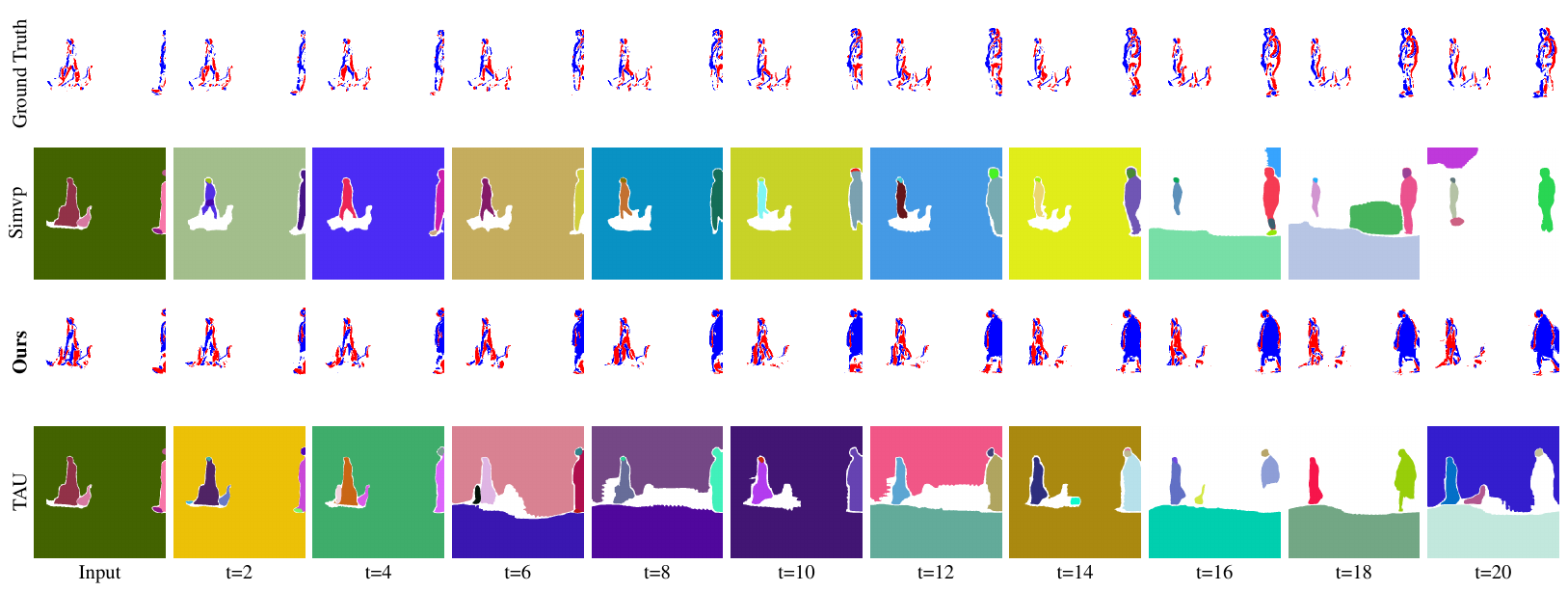}
    \end{subfigure}

    \vspace{0.00\textwidth}

    \begin{subfigure}[b]{1.0\textwidth}
        \centering
        \includegraphics[width=\textwidth]{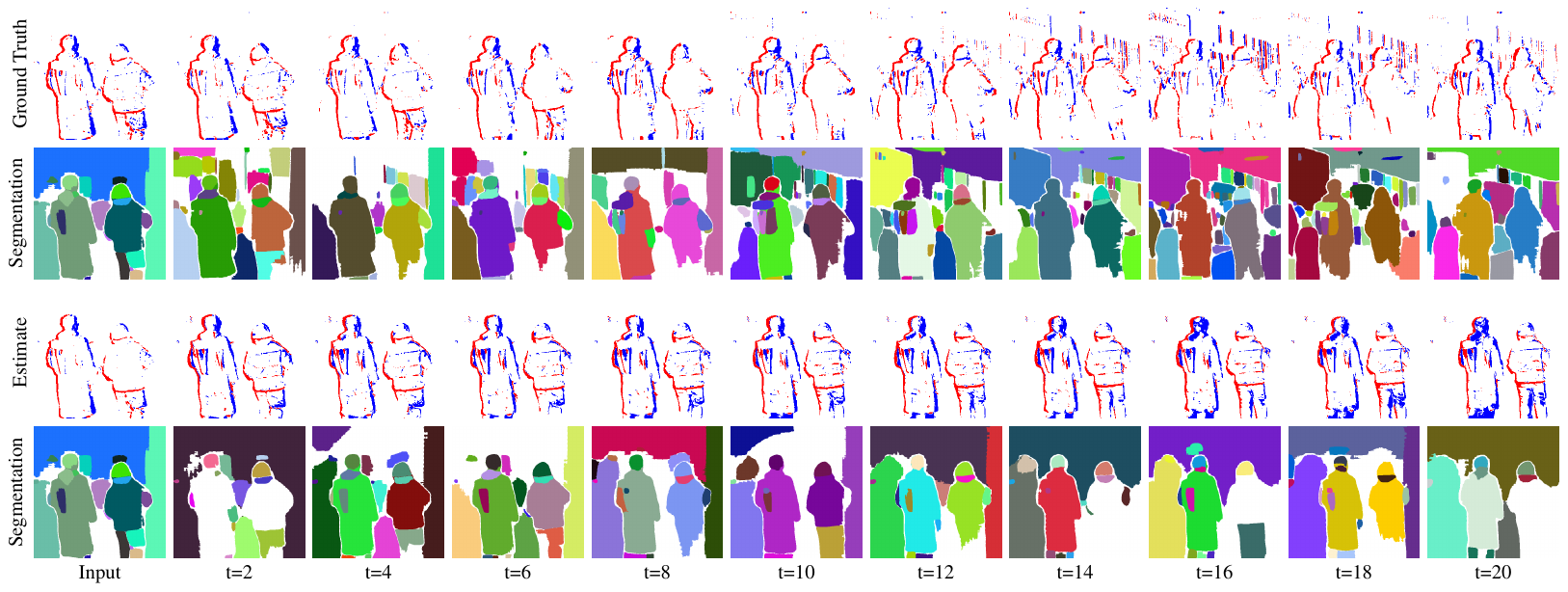}
    \end{subfigure}

    \vspace{0.00\textwidth}
    
    \begin{subfigure}[b]{1.0\textwidth}
        \centering
        \includegraphics[width=\textwidth]{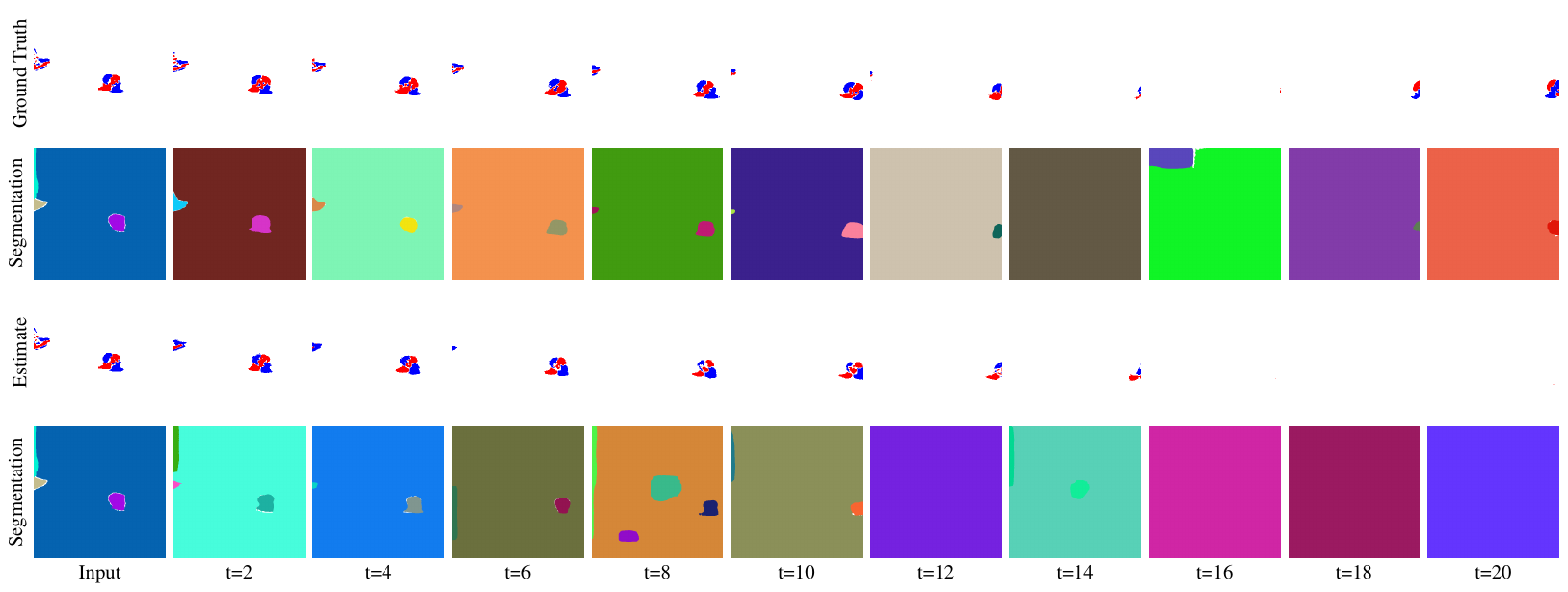}
    \end{subfigure}
    
    \vspace{0.01\textwidth}
    
    \begin{subfigure}[b]{1.0\textwidth}
        \centering
        \includegraphics[width=\textwidth]{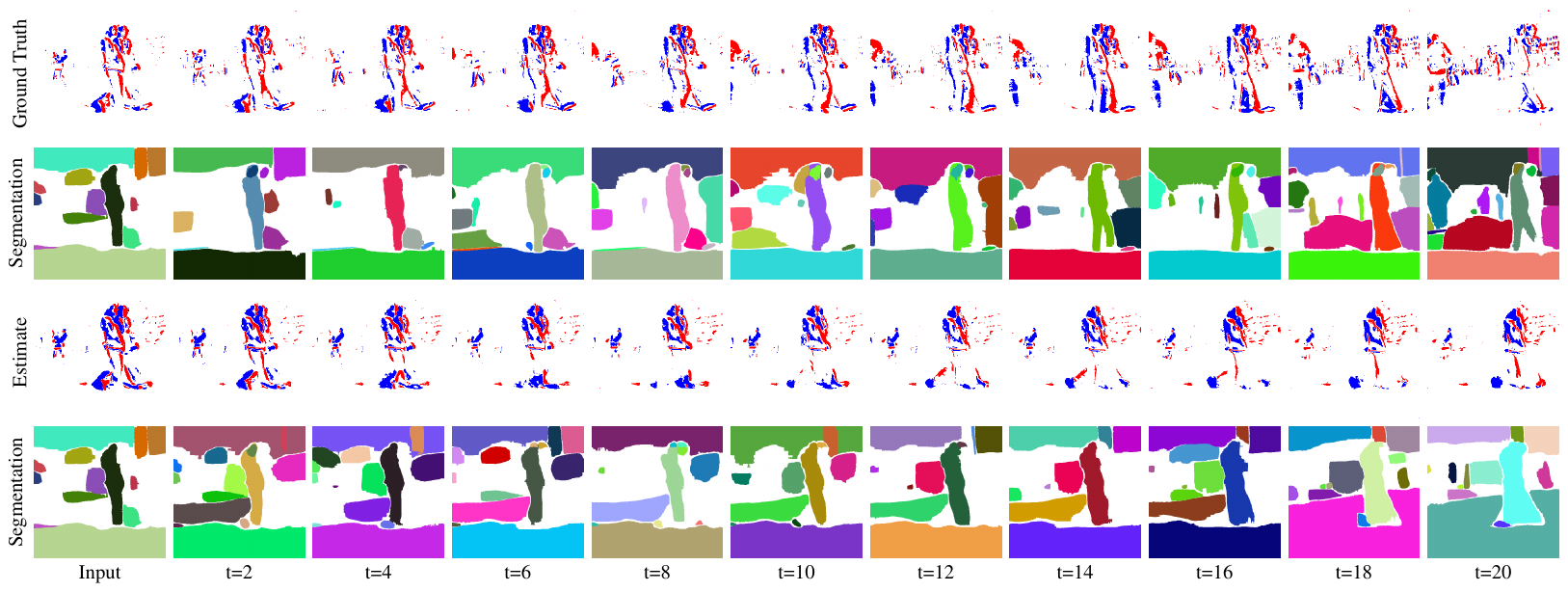}
    \end{subfigure}
    
    \caption{Additional results of Segmentation. The first row of each sequence represents the ground truth of the events, and the second row shows the segmentation results. The last two rows respectively display the results estimated by our method and the corresponding segmentation results.}
    \label{fig:segmentation}
    
\end{figure}

\begin{figure}[htbp]
    \centering
    \begin{subfigure}[b]{1.0\textwidth}
        \centering
        \includegraphics[width=\textwidth]{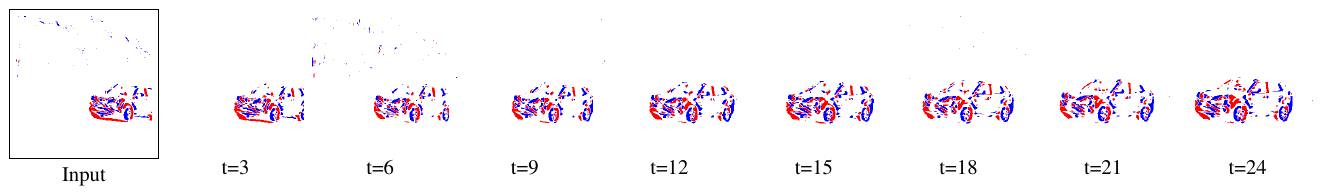}
        \caption{Ground Truth of the Car Sequence}
        \label{fig6:image1}
    \end{subfigure}

    \vspace{0.03\textwidth}

    \begin{subfigure}[b]{1.0\textwidth}
        \centering
        \includegraphics[width=\textwidth]{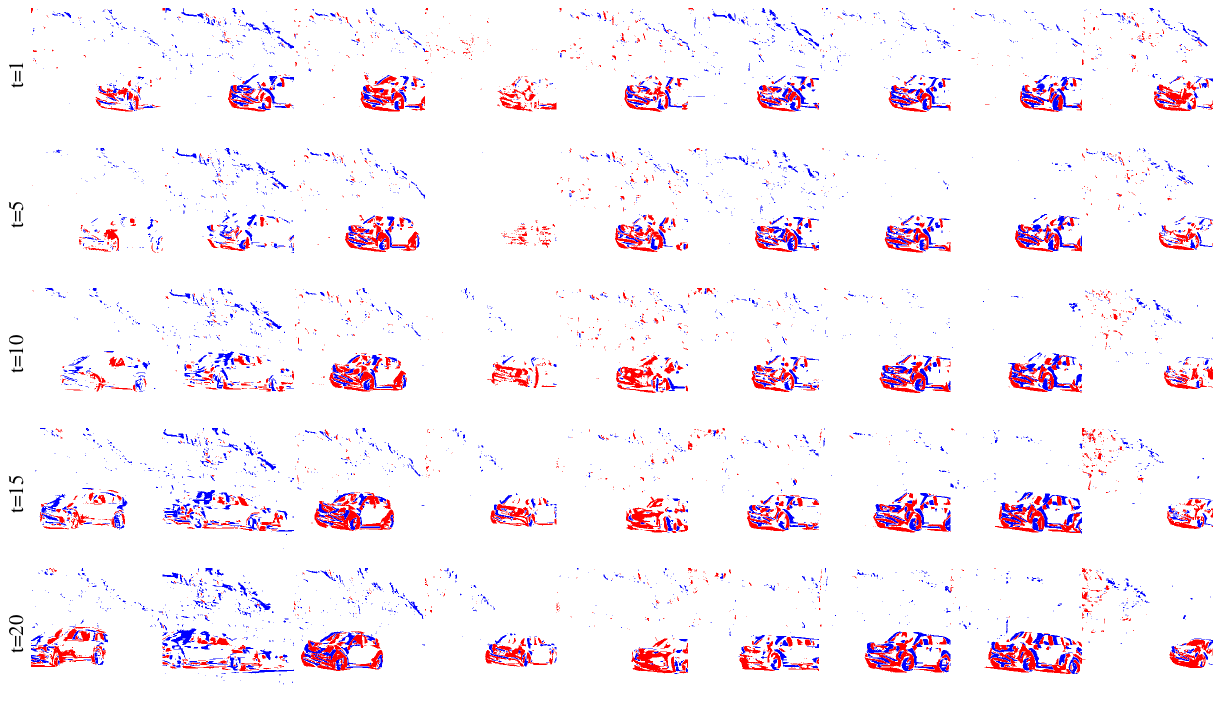}
        \caption{Visualization of the Pre-train model's Estimation Results(without Motion Alignment)}
        \label{fig6:image2}
    \end{subfigure}

    \vspace{0.03\textwidth}
    
    \begin{subfigure}[b]{1.0\textwidth}
        \centering
        \includegraphics[width=\textwidth]{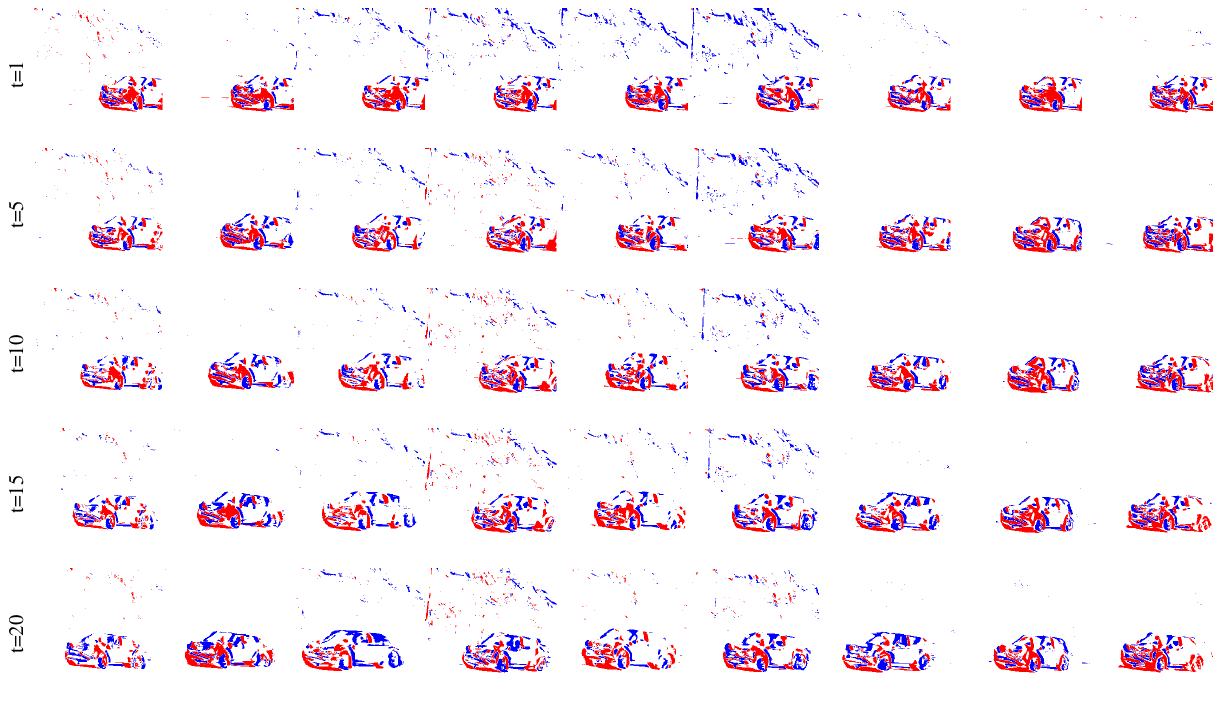}
        \caption{Visualization of the Results after Motion Alignment}
        \label{fig6:image3}
    \end{subfigure}
    
    \caption{Visualization results of Motion Alignment. Fig.~\ref{fig6:image1} shows the ground truth of the car sequence; Fig.~\ref{fig6:image2} presents the estimated results using a pre-trained model with 9 different random seeds, where several instances resulted in failures; Fig.\ref{fig6:image3} illustrates the motion alignment results generated by applying reinforcement learning, yielding more stable estimation.}
    \label{fig:rein_vis}
    
\end{figure}

\begin{figure}[htbp]
    \centering
    \begin{subfigure}{1.0\columnwidth}
        \centering
        \includegraphics[width=\columnwidth]{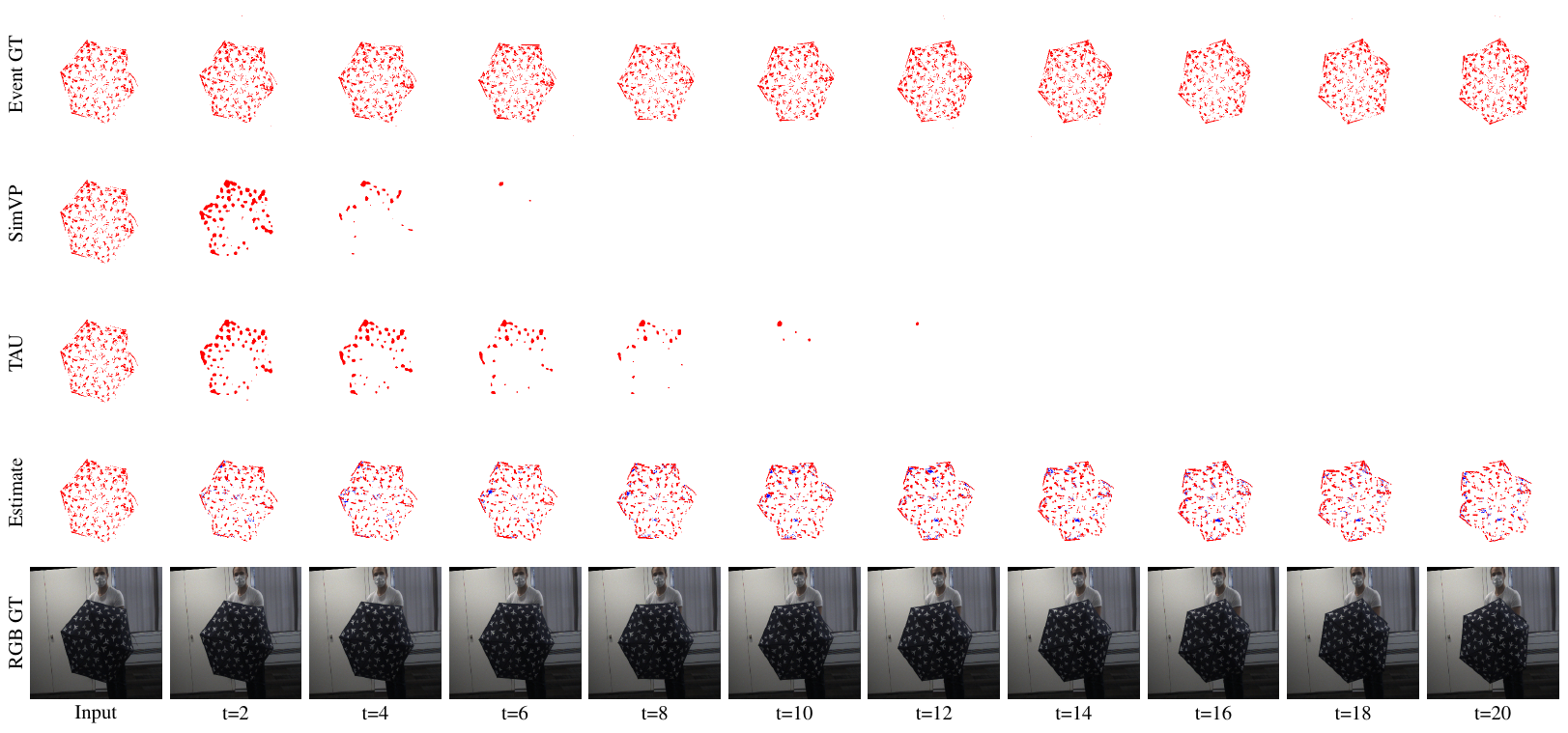}
        \caption{Rotating Umbrella.}
    \end{subfigure}

    \vspace{0.01\columnwidth}

    \begin{subfigure}{1.0\columnwidth}
        \centering
        \includegraphics[width=\columnwidth]{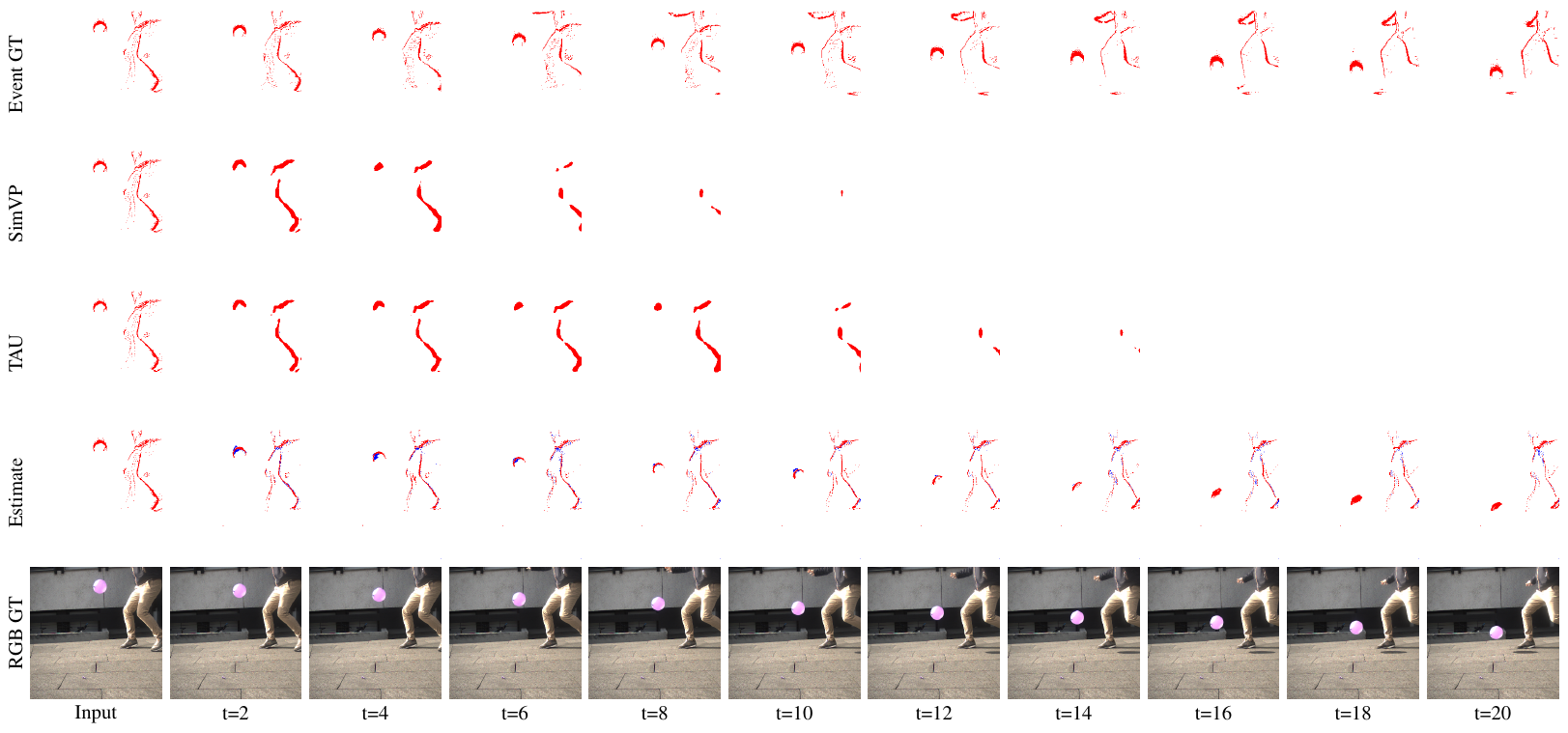}
        \caption{Falling Balloon.}
    \end{subfigure}

    \caption{Visualization of our method's prediction in hs-ergb dataset.}
    \label{fig:uzh}
    
\end{figure}

\begin{figure}[htbp]
    \centering
    \begin{subfigure}{1.0\columnwidth}
        \centering
        \includegraphics[width=\columnwidth]{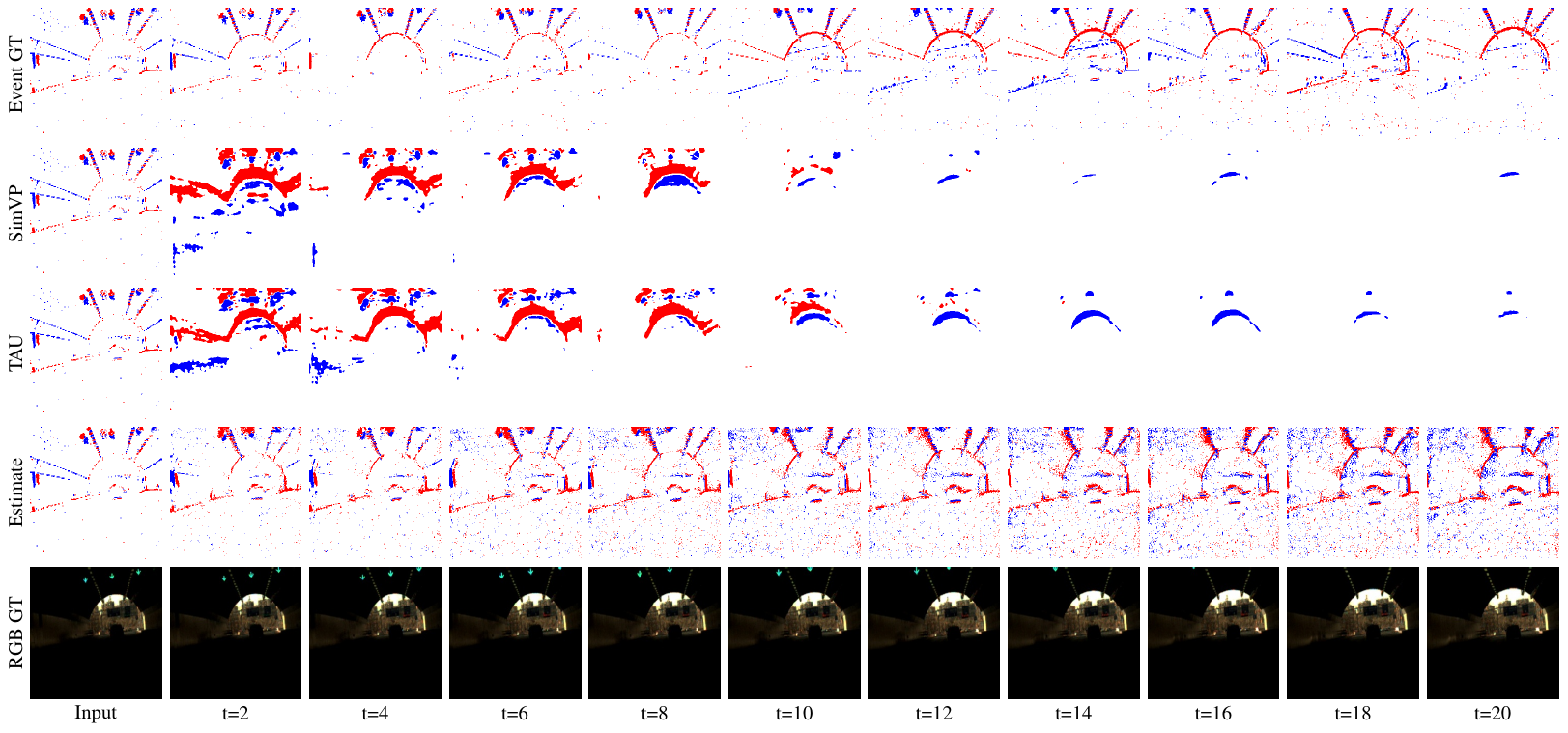}
        \caption{Poor exposure: The car driving out of the tunnel.}
    \end{subfigure}

    \vspace{0.01\columnwidth}

    \begin{subfigure}{1.0\columnwidth}
        \centering
        \includegraphics[width=\columnwidth]{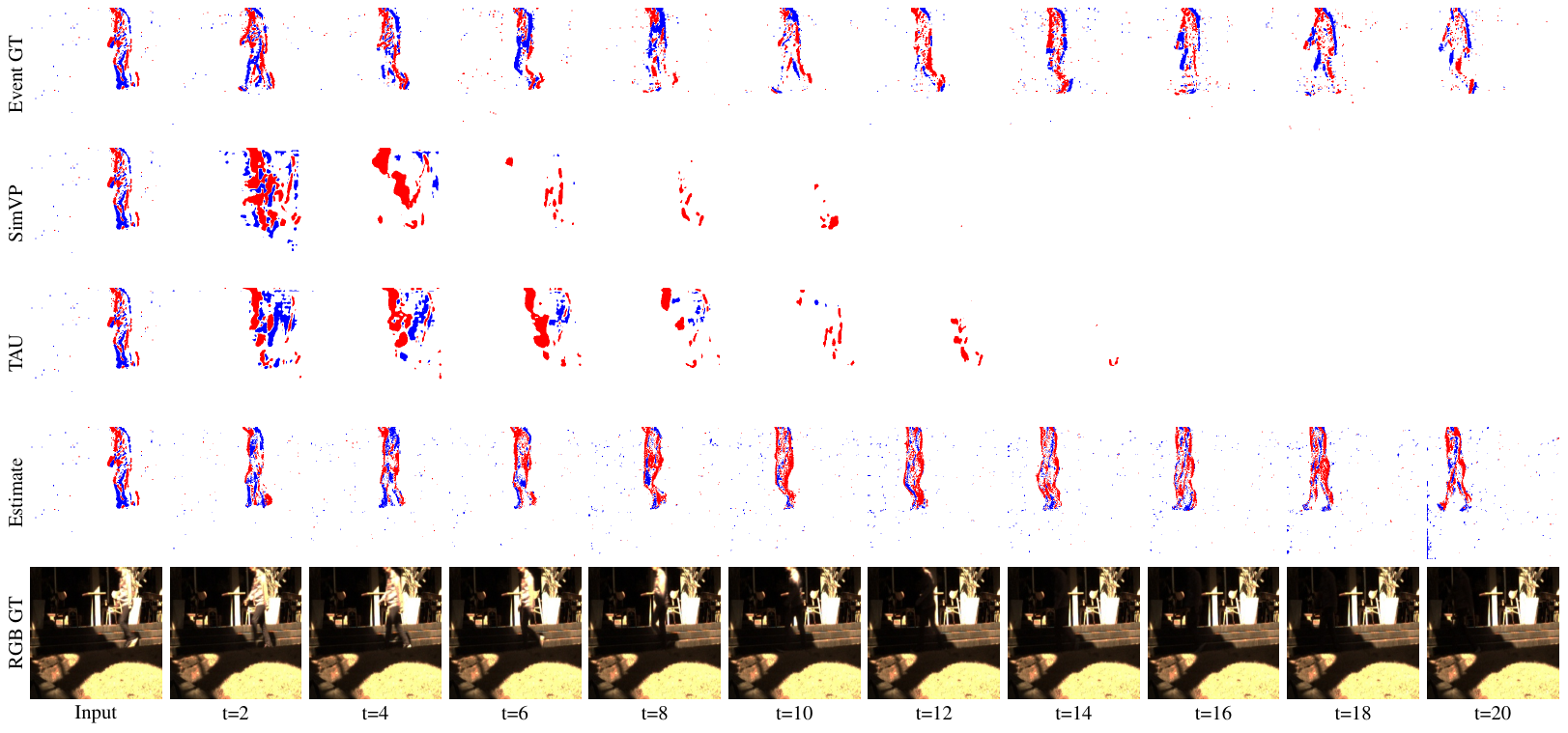}
        \caption{Occlusion: A person walking from a bright place to shadow.}
    \end{subfigure}

    \caption{Visualization of our method's prediction in various scenarios.}
    \label{fig:various}
    
\end{figure}

\section{Limitation}
\label{limit}
\subsection{Limitation scenarios}
\begin{figure}[t]
    \centering
    \begin{subfigure}{1.0\columnwidth}
        \centering
        \includegraphics[width=\columnwidth]{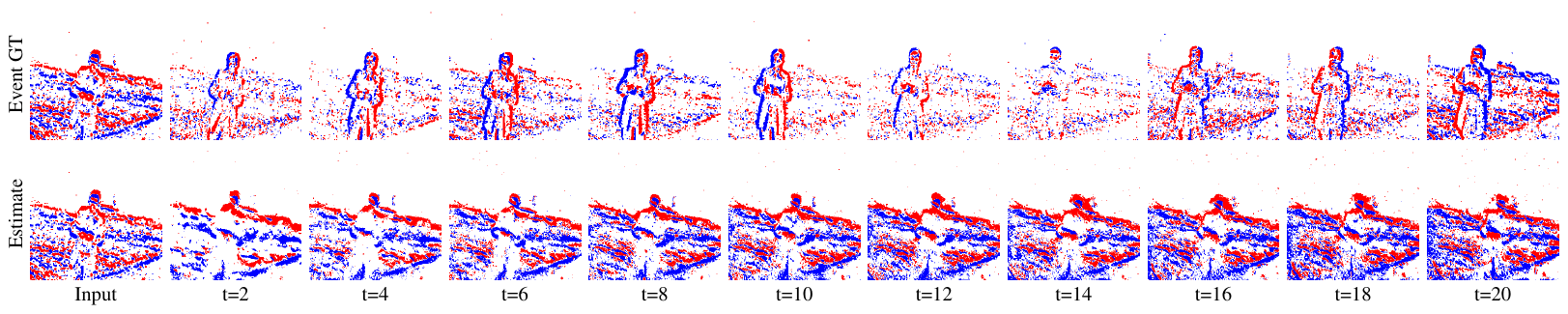}
        \caption{Shaky camera capturing a stationary person.}
        \label{fig4:complex}
    \end{subfigure}

    \vspace{0.01\columnwidth}

    \begin{subfigure}{1.0\columnwidth}
        \centering
        \includegraphics[width=\columnwidth]{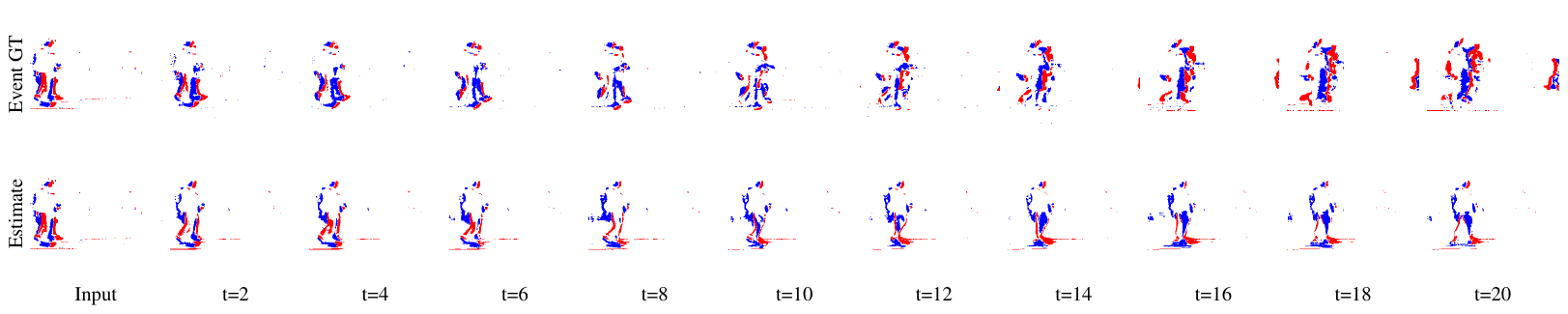}
        \caption{A person running out behind another person.}
        \label{fig4:sparse}
    \end{subfigure}

    \vspace{0.01\columnwidth}

    \begin{subfigure}{1.0\columnwidth}
        \centering
        \includegraphics[width=\columnwidth]{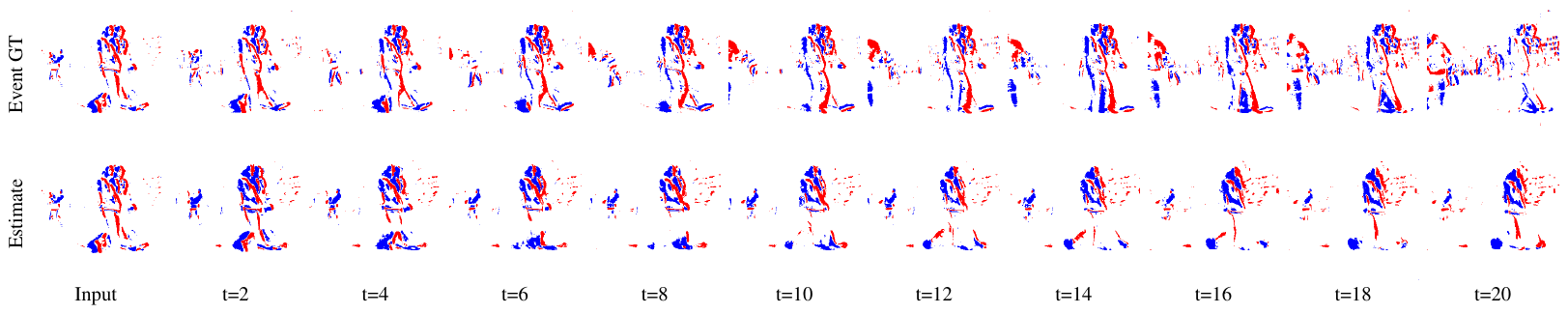}
        \caption{Two people walking side by side.}
        \label{fig4:overlap}
    \end{subfigure}
    
    \caption{Visualization of our method's prediction in severely degraded scenarios.}
    \label{fig:limited}
    
\end{figure}

Although our method achieves good results in most scenarios, it is still significantly limited in the following cases due to the sampling characteristics of event data: (1) \textbf{Complex background scenarios.} Event cameras may capture incomplete textures in certain situations, leading to poorer prediction performance, as shown in Fig.~\ref{fig4:complex}. Complex backgrounds can reduce the clarity of the target object, resulting in worse outcomes, especially in cases where the camera lens is shaking. (2) \textbf{Heavily overlapped object scenarios.} When objects overlap, their motion becomes quite complex, and due to the edge-focused characteristics of event cameras, understanding such motion is challenging, as shown in Fig.~\ref{fig4:sparse} and Fig.~\ref{fig4:overlap}. When people overlap, their footsteps often become chaotic, leading to less accurate predictions.

\subsection{Computational Resource \& Scalability}
The following Table.~\ref{table:computational} compares the computational resources of our method with SOTA methods. The powerful generative capability and high fidelity of diffusion models lead to the cost of substantial computational resource consumption. As shown in the following table, our parameter count and FLOPs significantly exceed those of traditional models. However, we believe this trade-off is necessary because of the powerful learning capability of large models in the real world. Taking the future motion estimation task as an example, our diffusion-based method significantly surpasses traditional methods in understanding and learning motion.

For a salable model size for different inference environments, there indeed are many works indicating that the diffusion model can be applied to quantization or other acceleration techniques for speeding up the inference process. We believe that with the advancement of hardware and acceleration techniques, the inference speed of diffusion models will be significantly improved in the near future.

\begin{table*}[t]
    \newcolumntype{C}{>{\centering\arraybackslash}p{1.2cm}}
    \caption{Comparison of methods in terms of parameters and FLOPs.}
    \label{table:computational}
    \centering
    \begin{tabular}{lcc}
        \toprule
        \textbf{Methods} & \textbf{Params (M)} & \textbf{FLOPs (G)} \\ 
        \midrule
        PredRNNv2        & 23.9                & 48.92              \\ 
        SimVP            & 58.0                & 60.61              \\ 
        TAU              & 44.7                & 92.50              \\ 
        \textbf{ours}    & 1521.0              & 693.92             \\ 
        \bottomrule
    \end{tabular}
\end{table*}

\end{document}